\documentclass[letterpaper]{article}
\usepackage{uai2018}
\usepackage[margin=1in]{geometry}

\usepackage{times}

\usepackage[T1]{fontenc}
\usepackage[latin1]{inputenc}
\usepackage[english]{babel}
\usepackage{amsopn}
\usepackage{url}
\usepackage{smallref}

\usepackage{cite}		

\usepackage{amsmath, amsfonts}
\usepackage{amssymb}
\usepackage{amsthm}
\usepackage[all]{xy}
\usepackage{cancel}
\usepackage{dsfont}
\usepackage{tikz}

\usetikzlibrary{arrows}
\usetikzlibrary{arrows.meta, calc}
\usetikzlibrary{shapes,positioning,decorations.markings}

\theoremstyle{plain}

\newtheorem{sa}{Theorem}[section]
\newtheorem{Thm}[sa]{Theorem}
\newtheorem{Lem}[sa]{Lemma}

\newtheorem{Cor}[sa]{Corollary}

\newtheorem{Def}[sa]{Definition}

\newtheorem{Pst}[sa]{Postulate}

\newtheorem{Rem}[sa]{Remark}

\newtheorem{Eg}[sa]{Example}

\newcommand{\R}{\mathbb{R}}

\newcommand{\Xcal}{\mathcal{X}}

\newcommand{\Scal}{\mathcal{S}}
\newcommand{\Lcal}{\mathcal{L}}

\newcommand{\sm}{\setminus}						
\newcommand{\ins}{\subseteq}

\newcommand*{\tut}[1][]{\mathrel{\tikz [baseline=-0.25ex,-, #1] \draw [#1] (0pt,0.5ex) -- (1.3em,0.5ex);}}

\newcommand*{\tuh}[1][]{\mathrel{\tikz [baseline=-0.25ex,-latex, #1] \draw [#1] (0pt,0.5ex) -- (1.3em,0.5ex);}}
\newcommand*{\hut}[1][]{\mathrel{\tikz [baseline=-0.25ex,latex-, #1] \draw [#1] (0pt,0.5ex) -- (1.3em,0.5ex);}}
\newcommand*{\huh}[1][]{\mathrel{\tikz [baseline=-0.25ex,latex-latex, #1] \draw [#1] (0pt,0.5ex) -- (1.3em,0.5ex);}}
\newcommand*{\ouo}[1][]{\mathrel{\tikz [baseline=-0.25ex,-, #1] \draw [
decoration={markings, mark=at position 0 with {\draw circle (1pt);}, 
mark=at position 1 with {\draw circle (1pt);}}, postaction={decorate},#1] (0pt,0.5ex) -- (1.3em,0.5ex);}}
\newcommand*{\tuo}[1][]{\mathrel{\tikz [baseline=-0.25ex,-, #1] \draw [
decoration={markings, mark=at position 1 with {\draw circle (1pt);}}, postaction={decorate},#1] (0pt,0.5ex) -- (1.3em,0.5ex);}}
\newcommand*{\huo}[1][]{\mathrel{\tikz [baseline=-0.25ex,latex-, #1] \draw [
decoration={markings, mark=at position 1 with {\draw circle (1pt);}}, postaction={decorate},#1] (0pt,0.5ex) -- (1.3em,0.5ex);}}
\newcommand*{\out}[1][]{\mathrel{\tikz [baseline=-0.25ex,-, #1] \draw [
decoration={markings, mark=at position 0 with {\draw circle (1pt);}}, postaction={decorate},#1] (0pt,0.5ex) -- (1.3em,0.5ex);}}
\newcommand*{\ouh}[1][]{\mathrel{\tikz [baseline=-0.25ex,-latex, #1] \draw [
decoration={markings, mark=at position 0 with {\draw circle (1pt);}}, postaction={decorate},#1] (0pt,0.5ex) -- (1.3em,0.5ex);}}

\newcommand*{\rars}{\begin{array}{c}\huh\\[-10pt]\tuh\end{array}}
\newcommand*{\lars}{\begin{array}{c}\huh\\[-10pt]\hut\end{array}}
\newcommand*{\allars}{\begin{array}{c}\huh\\[-10pt]\tuh\\[-10pt]\hut\\[-10pt]\tut\end{array}}
\newcommand*{\dars}{\begin{array}{c}\huh\\[-10pt]\tuh\\[-10pt]\hut\end{array}}

\renewcommand{\Pr}{\mathbb{P}} 		

\newcommand{\I}{\mathds{1}}
\newcommand{\Pa}{\mathrm{Pa}} 		
 		
\newcommand{\Ch}{\mathrm{Ch}} 		
\newcommand{\Anc}{\mathrm{Anc}} 		
 
\newcommand{\Desc}{\mathrm{Desc}}

\newcommand{\Sc}{\mathrm{Sc}}

\DeclareMathOperator*{\Indep}{{\,\perp\mkern-12mu\perp\,}}
\DeclareMathOperator*{\nIndep}{{\,\not\mkern-1mu\perp\mkern-12mu\perp\,}}
\DeclareMathOperator*{\given}{|}
\DeclareMathOperator*{\diag}{\mathrm{diag}}
\DeclareMathOperator{\doit}{do}
\newcommand{\moral}{\mathrm{mor}}

\newcommand{\ReLU}{\mathrm{ReLU}} 

\newcommand{\lp}{\left ( }
\newcommand{\rp}{\right ) }

\newcommand{\hyle}[2]{\emph{#2}}
\newcommand{\hyte}[2]{\emph{#2}}

\DeclareMathOperator*{\argmin}{arg\,min}

\newcommand{\xto}[1]{\stackrel{#1}{\to}}
\newcommand{\eRN}{\overline{\mathbb{R}}}
\newcommand{\RN}{\mathbb{R}}

\usepackage{times}

\usepackage{listings}
\lstdefinestyle{ASP}{
  belowcaptionskip=1\baselineskip,
  breaklines=true,
  frame=L,
  xleftmargin=\parindent,
  language=psl,
  showstringspaces=false,
  basicstyle=\script\ttfamily,
  keywordstyle=\bfseries\color{green!40!black},
  commentstyle=\itshape\color{purple!40!black},
  identifierstyle=\color{blue},
  stringstyle=\color{orange},
}

\title{Constraint-based Causal Discovery for 
Non-Linear Structural Causal Models with Cycles and Latent Confounders}

\author{ {\bf Patrick Forr\'e}\\
Informatics Institute\\University of Amsterdam\\The Netherlands\\
\url{p.d.forre@uva.nl}\\
\And
{\bf Joris M.~Mooij}\\
Informatics Institute\\University of Amsterdam\\The Netherlands\\
\url{j.m.mooij@uva.nl}\\
}

\begin{document}

\maketitle

\begin{abstract}
We address the problem of causal discovery from data, making use of the
recently proposed causal modeling framework of \emph{modular structural causal models (mSCM)}
to handle cycles, latent confounders and non-linearities.
We introduce \emph{$\sigma$-connection graphs ($\sigma$-CG)}, a new class of mixed graphs (containing undirected, bidirected and directed edges) with additional structure, and extend the concept of \emph{$\sigma$-separation}, the appropriate generalization of the well-known notion of d-separation in this setting, to apply to $\sigma$-CGs. We prove the closedness of $\sigma$-separation under marginalisation and conditioning and exploit this to implement a test of $\sigma$-separation on a $\sigma$-CG. 
This then leads us to the first causal discovery algorithm that can handle non-linear functional relations, latent confounders, cyclic causal relationships, and data from different (stochastic) perfect interventions.
As a proof of concept, we show on synthetic data how well the algorithm recovers features of the causal graph of modular structural causal models.
%
\end{abstract}




\section{INTRODUCTION}

Correlation does not imply causation. To go beyond spurious probabilistic associations and 
infer the asymmetric causal relations we need sufficiently powerful models. Structural causal models (SCMs), also known as structural equation models (SEMs), provide a popular modeling framework (see \cite{SGS00, Pearl09,PJS17,FM17}) that is up to this task.
Still, the problem of causal discovery from data is notoriously hard.
Theory and algorithms 
need to address several challenges like probabilistic settings, stability under interventions, combining observational and interventional data, 
latent confounders and marginalisation, selection bias and conditioning, faithfulness violations, cyclic causation like feedback loops and pairwise interactions, 
and non-linear functional relations in order to go beyond artificial simulation settings and become successful on real-world data. 

Several algorithms for causal discovery have been introduced over the years. For 
the acyclic case without latent confounders, numerous constraint-based \cite{Pearl09,SGS00} and score-based approaches \cite{HMC99,GeH94,CoH92} exist. More sophisticated constraint-based \cite{Pearl09,SGS00,CMH13,Zhang2008} and score-based approaches \cite{DER09, Eva15b, ER14, EvR10, CH13} can deal with latent confounders in the acyclic case. For the linear cyclic case, most algorithms assume no latent confounders \cite{Richardson96,RiS99,HEH10}, though some of the more recent ones allow for those \cite{HEJ14,Rothenhausler_15}. To the best of our knowledge, no algorithms have yet been proposed for the general non-linear cyclic case.

In this work we present a novel conditional independence constraint-based causal discovery algorithm that---up to the knowledge of the authors---is the first causal discovery algorithm that addresses most of the previously mentioned problems 
at once, 
notably non-linearities, cycles, latent confounders, and multiple interventional data sets, only excluding selection bias and faithfulness violations.

For this to work we build upon the theory of \emph{modular structural causal models (mSCM)} introduced in \cite{FM17}. 
mSCMs form a general and convenient class of structural causal models that can deal with cycles and latent confounders. The measure-theoretically rigorous presentation opens the door for general non-linear measurable functions and any kind of probability distributions (e.g.\ mixtures of discrete or continuous ones). mSCM are provably closed under any combination of perfect interventions and marginalisations (see \cite{FM17}).

Unfortunately, it is known that the direct generalization of the \emph{d-separation criterion} (also called m- or m$^*$-separation, see \cite{Pearl86b,VerPea90a,Pearl09,Eva15,Richardson03}), which relates the conditional independencies of the model to its underlying graphical structure, does not apply in general if the structural equations are \emph{non-linear} and the graph contains \emph{cycles} (see \cite{Spirtes95, FM17} or example \ref{the-example}).

Luckily, one key property of mSCMs is that the variables of the mSCM always entail the conditional independences implied by 
\emph{$\sigma$-separation}, a non-naive generalization of the d/m/m$^*$-separation (see \cite{FM17}), which also works in the presence of cycles, non-linearities and latent confounders, and reduces to d-separation in the acyclic case.

To prove the $\sigma$-separation criterion, the authors of \cite{FM17} have constructed an extensive theory for directed graphs with hyperedges. 
As a first contribution in this paper we give a simplified but equivalent definition of mSCMs plainly in terms of directed graphs and prove the $\sigma$-separation criterion directly under weaker assumptions.

As a second contribution we extend the definition of $\sigma$-separation to mixed graphs (including also bi- and undirected edges) by introducing additional structure.
We will refer to this class of mixed graphs as \emph{$\sigma$-connection graphs ($\sigma$-CG)}, since they are 
inspired by the d-connection graphs introduced by \cite{HEJ14}. 
We prove that $\sigma$-CGs and $\sigma$-separation are closed under marginalisation and conditioning,
in analogy with the d-connection graphs (d-CG) from \cite{HEJ14}.


The work of \cite{HEJ14} provides an elegant approach to causal discovery using a weighted SAT solver to find the causal graph that is most compatible with (weighted) conditional independences in the data, encoding the notion of d-separation into 
\emph{answer set programming (ASP)}, a declarative programming language with an expressive syntax for implementing discrete or integer optimization problems.
Our third contribution is to adapt the approach of \cite{HEJ14} by replacing d-separation by $\sigma$-separation and d-CGs by $\sigma$-CGs. The results mentioned above will then ensure all the needed properties to make the adapted algorithm of \cite{HEJ14} applicable to general mSCMs, i.e., to non-linear causal models with cycles and latent confounders, under the additional assumptions of \emph{no selection bias} and of \emph{$\sigma$-faithfulness}.

Finally, as a proof of concept, we will show the effectiveness of our proposed algorithm in 
recovering features of the causal graphs of mSCMs from simulated data. 

\section{THEORY}

\subsection{MODULAR STRUCTURAL CAUSAL MODELS}

\begin{figure}[h]
  \includegraphics[scale=0.43]{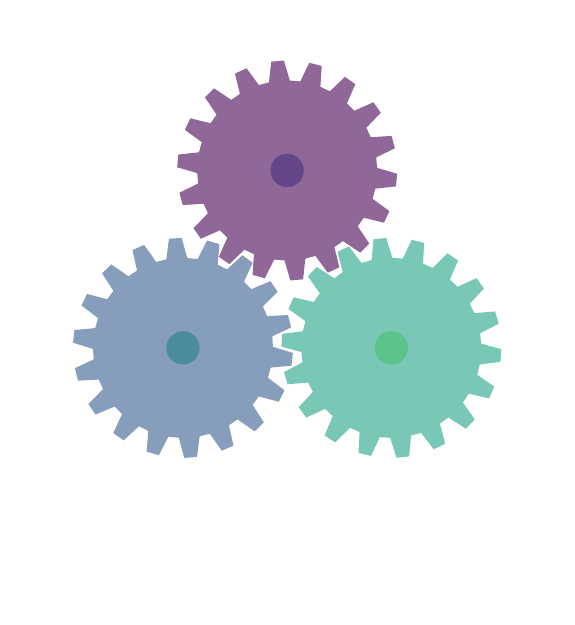}
	\includegraphics[scale=0.43]{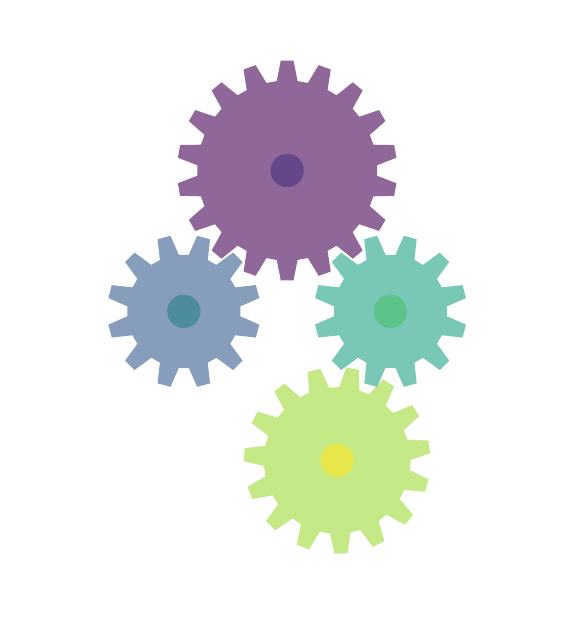}
	\includegraphics[scale=0.43]{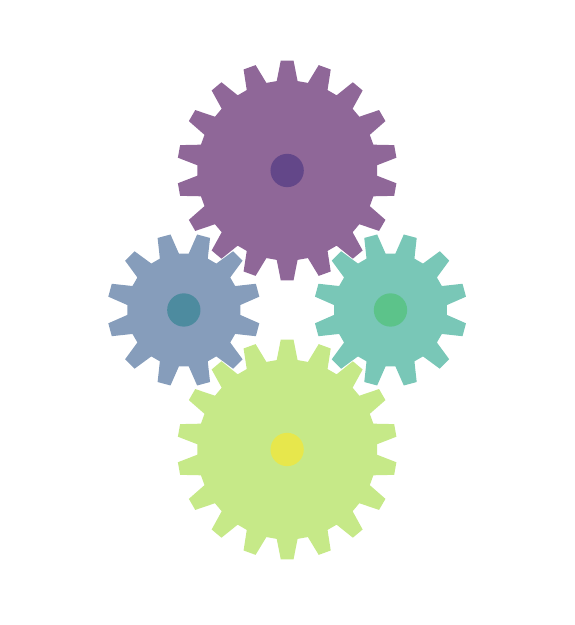}
	\caption{Gear analogy. Left: A cyclic mechanism that is locally compatible but not globally.
	 Center: For acyclic mechanisms local compatibility implies global compatibility. 
	Right: A cyclic mechanism that is locally and globally compatible. 
	This shows that the assumption of global compatibility is needed when cycles are present.	}
	 \label{fig:gears}
\end{figure}

Structural causal or equation models (SCM/SEM) usually start with a set of variables $(X_v)_{v \in V}$ attached to a graph $G$ that satisfy (or are even defined by) equations of the form:
\[ X_v = g_{\{v\}}(X_{\Pa^G(v)},E_v), \]
with a function $g_{\{v\}}$ and noise variable $E_v$ attached to each node $v \in V$. 
Here ${\Pa^G(v)}$ denotes the set of (direct causal) parents of $v$.
In \emph{linear} models the functions $g_{\{v\}}$ are linear functions, in \emph{acyclic} models the nodes of $V$ form an acyclic graph $G$, and under \emph{causal sufficiency} the variables $E_v$ are independent (i.e.\ ``no latent confounders''). 
The functions $g_{\{v\}}$ are usually interpreted as \emph{local causal mechanisms} that produce the values of $X_v$ from the values of $X_{\Pa^G(v)}$ and $E_v$. These local mechanisms $g_{\{v\}}$ are---in the causal setting---assumed to be stable even when one intervenes upon some of the variables, i.e.\ one makes a causal \emph{local compatibility} assumption.
One important observation now is that one can also consider 
the \emph{global mechanism} $g$ that maps the values of the latent variables $(E_v)_{v \in V}$ to the values of the observed variables $(X_v)_{v \in V}$.
The assumption of acyclicity or invertible linearity will then guarantee the \emph{global compatibility} of all these mechanisms $g_{\{v\}}$ and $g$.
However, if we now abstain from assuming acyclicity or linearity, the global compatibility does not follow from the local compatibility anymore (see figure \ref{fig:gears}). So in a general consistent causal setting this needs to be \emph{guaranteed or assumed}.


The definition of modular structural causal models (mSCM) and the mentioned list of desirable properties basically follow directly from \emph{causal} postulates:

\begin{Pst}[Causal postulates]$ $
\label{two-assumptions}
The observed world appears as the projection of an extended world such that:
\begin{enumerate}
  \setlength\itemsep{0em}
  \item  All latent and observed variables in this extended world are causally linked  by a directed graph.
 \item Every subsystem of this extended world can be expressed as the joint effect of its joint direct causes.
 \item All these mechanisms are globally compatible.
\end{enumerate}
\end{Pst}

Special subsystems of interest are the loops of a graph. 

\begin{Def}[Loops]
Let $G=(V,E)$ be a directed graph (with or without cycles).
\begin{enumerate}
  \setlength\itemsep{0em}
  \item A \hyte{loop}{loop} of $G$ is a set of nodes $S \ins V$ 
	such that for every two nodes $v_1,v_2 \in S$ there are two directed paths $v_1 \tuh \cdots \tuh v_2$ and  $v_2 \tuh \cdots \tuh v_1$ in $G$ with all the intermediate nodes also in $S$ (if any). The single element sets $S=\{v\}$ are also considered as loops.
\item The \hyte{Sc}{strongly connected component} of $v$ in $G$ is defined to be:
 \[ \Sc^G(v):= \Anc^G(v) \cap \Desc^G(v), \] 
the set of nodes that are both ancestors and descendants of $v$ (including $v$ itself).
\item Let $\Lcal(G):=\{ S \ins G \;|\, S \text{ a loop of } G  \}$ be the \hyte{loopset}{loop set of $G$}.
\end{enumerate}
\end{Def}

\begin{Rem}
Note that the loop set $\Lcal(G)$ contains all single element loops $\{v\} \in \Lcal(G)$, $v \in V$, as the smallest loops and all strongly connected components $\Sc^G(v) \in \Lcal(G)$, $v \in V$, as the largest loops, but also all non-trivial intermediate loops $S$ with $\{v\} \subsetneq S \subsetneq \Sc^G(v)$ inside the strongly connected components (if existent). If $G$ is acyclic then $\Lcal(G)$ only consists of the single element loops: $\Lcal(G)=\{\{v\} \,|\, v \in V\}$.
\end{Rem}

The definition of mSCM is made in such a way that it will automatically incorporate the causal postulates \ref{two-assumptions}. 
In the following, all spaces are meant to be equipped with $\sigma$-algebras, forming standard measurable spaces, and all maps to be measurable.

\begin{Def}[Modular Structural Causal Model, \cite{FM17}]
\label{mSCM-def}
A \hyte{mSCM}{modular structural causal model (mSCM)} by definition consists of:
\begin{enumerate}
\setlength\itemsep{0em}
\item a set of nodes $V^+=U \dot \cup V$, where elements of $V$ correspond to observed variables and elements of $U$ to latent variables,
\item an observation/latent space $\Xcal_v$ for every $v \in V^+$, $\Xcal:=\prod_{v \in V^+}\Xcal_v$,
\item a product probability measure $\Pr:=\Pr_U=\otimes_{u \in U} \Pr_u$ on the latent space $\prod_{u \in U} \Xcal_u$,\footnote{The assumption of independence of the noise variables here is not to be confused with causal sufficiency. The noise variables here might have two or more child nodes and thus can play the role of latent confounders. The independence assumption here also does not restrict the model class. If they were dependent we would just consider them as one variable and use a different graph that encoded this.}
 \label{3}
\item a directed graph structure $G^+=(V^+, E^+)$
 with the properties:\footnote{Even though we allow for selfloops in the directed graph $G^+$ we note that the causal mechanisms $g_S$ will depend only on $\Pa^{G^+}(S)\sm S$, removing the self-dependence on the functional level. Otherwise, the functions $g_S$ would not hold up to a direct interventional interpretation and one would want to replace them with functions that do.}
	\begin{enumerate}
	\setlength\itemsep{0em}
	  \item $V = \Ch^{G^+}(U)$,
		\item $\Pa^{G^+}(U)=\emptyset$,\footnote{This assumption is only necessary to give the mSCM a ``reduced/summarized'' form. In practice one could allow for more latent variables and more complex latent structure.}
		\item $\Ch^{G^+}(u_1) \nsubseteq \Ch^{G^+}(u_2)$ for every two distinct $u_1,u_2 \in U$,\footnotemark[3] 
	\end{enumerate}
	where $\Ch^{G^+}$ and $\Pa^{G^+}$ stand for children and parents in $G^+$, resp.,
\item a system of structural equations $g=(g_S)_{\substack{S \in \Lcal(G^+)\\ S \ins V}}$: 
 \[ g_S: \; \prod_{v \in \Pa^{G^+}(S)\sm S}\Xcal_v \to \prod_{v \in S}\Xcal_v,\footnotemark[2] \footnote{Note that the index set runs over all ``observable loops'' $S \ins V$, $S \in \Lcal(G^+)$, which contains the usual single element sets $S=\{v\}$, which relate to the usual mechanisms $g_{\{v\}}$.}\]
that satisfy the following \emph{global compatibility} conditions: For every nested pair of loops $S' \ins S \ins V$ of $G^+$ and every element
$x_{\Pa^{G^+}(S) \cup S} \in \prod_{ v\in \Pa^{G^+}(S) \cup S} \Xcal_v$ we have the implication:
\[\begin{array}{rcrcl} 
 && g_S(x_{\Pa^{G^+}(S) \sm S}) &=& x_S \\
 &\implies& g_{S'}(x_{\Pa^{G^+}(S') \sm S'})&=&x_{S'},  \end{array}\]
where $x_{\Pa^{G^+}(S') \sm S'}$ and $x_{S'}$ denote the corresponding components of $x_{\Pa^{G^+}(S) \cup S}$.
\end{enumerate}
The \hyle{mSCM}{mSCM} can be summarized by the tuple $M=(G^+,\Xcal,\Pr,g)$.
\end{Def}

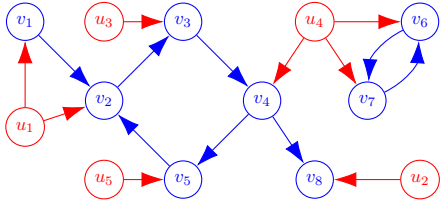
\begin{figure}[h]
\centering
\begin{tikzpicture}[scale=.7, transform shape]
\tikzstyle{every node} = [draw,shape=circle,color=blue]
\node (v1) at (0,0) {$v_1$};
\node (v2) at (1.5,-1.5) {$v_2$};
\node (v3) at (3,0) {$v_3$};
\node (v4) at (4.5,-1.5) {$v_4$};
\node (v5) at (3,-3) {$v_5$};
\node (v6) at (7.5,0) {$v_6$};
\node (v7) at (6.5,-1.5) {$v_7$};
\node (v8) at (5.5,-3) {$v_8$};
\tikzstyle{every node} = [draw,shape=circle, color=red]
\node (u7) at (5.5,0) {$u_4$};
\node (u1) at (0,-2) {$u_{1}$};
\node (u5) at (1.5,-3) {$u_5$};
\node (u3) at (1.5,0) {$u_3$};
\node (u8) at (7.5,-3) {$u_2$};
\draw[-{Latex[length=3mm, width=2mm]}, bend right, color=blue] (v7) to (v6);
\draw[-{Latex[length=3mm, width=2mm]}, bend right, color=blue] (v6) to (v7);
\foreach \from/\to in {v1/v2, v2/v3, v3/v4, v4/v5, v5/v2, v4/v8}
\draw[-{Latex[length=3mm, width=2mm]},color=blue] (\from) -- (\to);
\foreach \from/\to in {u7/v4, u7/v6, u7/v7, u1/v2, u3/v3, u1/v1, u5/v5, u8/v8}
\draw[-{Latex[length=3mm, width=2mm]},color=red] (\from) -- (\to);
\end{tikzpicture}
\caption{The graph $G^+$ of a modular structural causal model (mSCM). The observed variables $v_i \in V$ are in blue and the latent variables $u_j\in U$ are in red.
We have the four observed strongly connected components: $\{v_1\}$, $\{v_2,v_3,v_4,v_5\}$, $\{v_6,v_7\}$, $\{v_8\}$.
}
 \label{fig:mSCM-graph}
\end{figure}


\begin{Rem}
Given the mechanisms attached to the nodes $g_{\{v\}}$ the existence (and compatibility) of the other mechanisms $g_S$ for non-trivial loops $S$ can be guaranteed under certain conditions, e.g.\ trivially in the acyclic case, or if every cycle is contractive (see subsection \ref{mSCM-Lipschitz}), or more generally if the cycles are ``uniquely solvable'' (see \cite{FM17,Bongers++_1611.06221v2}).
\end{Rem}

We are now going to define the actual random variables $(X_u)_{u \in V^+}$ attached to any mSCM.

\begin{Rem}
Let $M=(G^+,\Xcal,\Pr,g)$ be a \hyle{mSCM}{mSCM} with $G^+=(U \dot \cup V, E^+)$.
\begin{enumerate}
\setlength\itemsep{0em}
\item The latent variables $(X_u)_{u \in U}$ 
are given by the canonical projections $E_u:\; \prod_{u' \in U} \Xcal_{u'} \to \Xcal_u$ and are jointly $\Pr$-independent (by \ref{3}). Sometimes we will write $(E_u)_{u \in U}$ instead of $(X_u)_{u \in U}$ to stress their interpretation as error/noise variables.
	\item The observed variables $(X_v)_{v \in V}$ are inductively defined by:
	 \[X_v := g_{S,v}\big((X_w)_{w \in \Pa^{G^+}(S)\sm S}\big),\] 
	where $S:=\Sc^{G^+}(v):=\Anc^{G^+}(v) \cap \Desc^{G^+}(v)$ and 
	where the second index $v$ refers to the $v$-component of $g_S$. Note that the inductive definition is possible because when ``aggregating'' each of the biggest cycles $\Sc^{G^+}(v)$ into one node then only an acyclic graph is left, which can be totally ordered.
		\item By the compatibility condition for $g$ we then have that for every $S \in \Lcal(G^+)$ with $S \ins V$ the following equality holds:
	\[X_S = g_{S}(X_{\Pa^{G^+}(S)\sm S}), \]
	where we put $X_A:=(X_v)_{v \in A}$ for subsets $A$.
\end{enumerate}	
\end{Rem}

As a consequence of the convenient definition \ref{mSCM-def} all the following desirable constructions 
(like marginalisations and interventions) 
are easily seen to be well-defined (for proofs see \cite{FM17}).
Note that already defining these constructions was a key challenge in the theory of causal models in the presence of cycles (see \cite{FM17,Bongers++_1611.06221v2}, a.o.).

\begin{Def}
\label{mSCM-marg-intv}
Let $M=(G^+,\Xcal,\Pr,g)$ be a \hyle{mSCM}{mSCM} with $G^+=(U \dot \cup V, E^+)$.
\begin{enumerate}	
	\item By plugging the functions $g_S$ into each other we can define the \emph{marginalised} \hyle{mSCM}{mSCM} $M'$ w.r.t.\ a subset $W \ins V$.
	 For example, when marginalizing out $W=\{w\}$ we can define (for the non-trivial case $w \in  \Pa^{G^+}(S) \sm S$):
	  \[\begin{array}{l} g_{S',v}(x_{\Pa^{G'}(S') \sm S'}) :=\\
		g_{S,v}(x_{\Pa^{G^+}(S) \sm (S \cup \{w\})}, g_{\{w\}}(x_{\Pa^{G^+}(w)\sm\{w\} })),  
		\end{array}\]
		where $G'$ is the marginalised graph of $G^+$, $S' \ins V'$ is any loop of $G'$ and $S$ the corresponding induced loop in $G^+$. 
	\item For a subset $I \ins V$ and a value $x_I \in \prod_{v \in I} \Xcal_v$ we define the \emph{intervened} graph $G'$ by removing all the edges from parents of $I$ to $I$. We put $X_u^{\doit(x_I)} := X_u$ for $u \in U$ and $X_I^{\doit(x_I)}:=x_I$ and inductively ($S:=\Sc^{G'}(v)$):
	\[X_v^{\doit(x_I)} := g_{S,v}(X^{\doit(x_I)}_{\Pa^{G'}(S)\sm S}). \]
	By selecting all functions $g_S$ where $S$ is still a loop in the intervened graph $G'$ we get the \emph{post-interventional} \hyle{mSCM}{mSCM} $M'$. These constructions give us all interventional distributions, e.g. (cf. \cite{Pearl09}):
	$$\Pr( X_A | \doit(x_I),X_B) := \Pr( X_A^{\doit(x_I)}|X_B^{\doit(x_I)} ).  $$
	Instead of fixing $X_I^{\doit(x_I)}$ to a value $x_I$ we could also specify a distribution $\Pr_I'$ for it (``randomization''). In this way we define stochastic interventions $\doit(\xi_I)$ with an independent random variable $\xi_I$ taking values in $\mathcal{X}_I$ and get a $\Pr^{\doit(I)}$ similarly. 
\end{enumerate}
\end{Def}


\subsection{$\Sigma$-SEPARATION IN MSCMS AND $\Sigma$-CONNECTION GRAPHS}

We now introduce $\sigma$-separation as a generalization of d-separation directly on the level of mixed graphs. To make the definition stable under marginalisation and conditioning we need to carry extra structure. The resulting graphs will be called $\sigma$-connection graphs ($\sigma$-CG), where the name is inspired by \cite{HEJ14}. An example that shall clarify the difference between d- and $\sigma$-separation is given later in figure \ref{fig:mSCM-graph} and table \ref{tab:mSCM-indeps}.

\begin{Def}[$\sigma$-Connection Graphs ($\sigma$-CG)]
  A \hyte{s-cg}{$\sigma$-connection graph ($\sigma$-CG)} is a mixed graph $G$ with a set of nodes $V$ and directed ($\tuh$), undirected ($\tut$) and bidirected ($\huh$) edges, together with an equivalence relation $\sim_\sigma$ on $V$ that has the property that every equivalence class $\sigma(v)$, $v \in V$, is a loop in the underlying directed graph structure: $\sigma(v) \in \Lcal(G)$.
  Undirected self-loops ($v \tut v$) are allowed, (bi)-directed self-loops ($v \tuh v$, $v \huh v$) are not.
\end{Def}
In particular, every node is assigned to a unique fixed loop $\sigma(v)$ in $G$ with $v \in \sigma(v)$ and two of such loops $\sigma(v_1)$, $\sigma(v_2)$ are either identical or disjoint. The reason for why we need such structure is illustrated in figure \ref{fig:cCG-acyclic-cyclic}.

\begin{Def}[$\sigma$-Open Path in a $\sigma$-CG]
\label{s-open-path}
Let $G$ be a $\sigma$-CG with set of nodes $V$ and $Z \ins V$ a subset.
Consider a path $\pi$ in $G$ with $n \ge 1$ nodes: 
\[ v_1 \allars  \cdots \allars v_n.\footnote{\label{fn2}The stacked edges are meant to be read as an ``OR'' at each place independently. We also allow for repeated nodes in the paths.} \]
The path will be called \emph{$Z$-$\sigma$-open} if:
 \begin{enumerate}
\setlength\itemsep{0em}
   \item 
	the \emph{endnodes} $v_1, v_n \notin Z$, and
   \item every triple of adjacent nodes in $\pi$ that is of the form:
	\begin{enumerate}
	\setlength\itemsep{0em}
	  \item \emph{collider}:
    \[v_{i-1} \rars v_i \lars v_{i+1},\]
		satisfies $v_i \in Z$,
	 \item \emph{(non-collider) left chain}:
	  \[v_{i-1} \hut v_i \lars v_{i+1}, \] 
	  satisfies  $v_i \notin Z$ or $v_i \in Z \cap \sigma(v_{i-1})$,
	 \item \emph{(non-collider) right chain}:
		\[v_{i-1}\rars v_i \tuh v_{i+1},\] 
		satisfies $v_i \notin Z$ or  $v_i \in Z \cap \sigma(v_{i+1})$,
		\item \emph{(non-collider) fork}:
	    \[v_{i-1} \hut v_i \tuh v_{i+1},\]
			 satisfies $v_i \notin Z$ or $v_i \in Z \cap \sigma(v_{i-1}) \cap \sigma(v_{i+1})$,
	 \item \emph{(non-collider) with undirected edge}:
	\[
	\begin{array}{c} 
	v_{i-1} \begin{array}{c}\tut\end{array} v_i \begin{array}{c}\tuh\\[-10pt]\hut\\[-10pt]\huh\\[-10pt]\tut\end{array} v_{i+1}, \\  
	v_{i-1}\begin{array}{c}\tuh\\[-10pt]\hut\\[-10pt]\huh\\[-10pt]\tut\end{array} v_i \begin{array}{c}\tut\end{array} v_{i+1},
	 \end{array} 	\] 
	satisfies $v_i \notin Z$.
	\end{enumerate}
	\end{enumerate}
\end{Def}

The difference between $\sigma$- and d-separation lies in the additional conditions involving $Z \cap \sigma(v_{i\pm 1})$. The intuition behind them is that the dependence structure \emph{inside} a loop $\sigma(v_i)$ is so strong that non-colliders can only be blocked by conditioning if an edge is pointing out of the loop (see example  \ref{the-example} and table \ref{tab:mSCM-indeps}).

Similar to d-separation we can now define $\sigma$-separation in a $\sigma$-CG.

\begin{Def}[$\sigma$-Separation in a $\sigma$-CG]
\label{s-sep-def}
Let $G$ be a $\sigma$-CG with set of nodes $V$. 
 Let $X,Y,Z \ins V$ be subsets. 
\begin{enumerate}
\setlength\itemsep{0em}
\item We say that $X$ and $Y$ are \hyte{ns-sep}{$\sigma$-connected by $Z$ or not $\sigma$-separated by $Z$} if there exists a path $\pi$ (with some $n \ge 1$ nodes) in $G$ with one endnode in $X$ and one endnode in $Y$ that is $Z$-$\sigma$-open. In symbols this statement will be written as follows:
\[ X \nIndep^\sigma_G Y \given Z.\]
\item Otherwise, we will say that $X$ and $Y$ are \hyte{s-sep}{$\sigma$-separated} by $Z$ and write:
\[ X \Indep^\sigma_G Y \given Z.\]
\end{enumerate}
\end{Def}

\begin{Rem}
\begin{enumerate}
\setlength\itemsep{0em}
\item The \emph{finest/trivial} $\sigma$-CG structure of a mixed graph $G$ is given by $\sigma(v):=\{v\}$ for all $v \in V$. In this way $\sigma$-separation in $G$ coincides with the  usual notion of d-separation in a d-connection graph (d-CG) $G$ (see \cite{HEJ14}). We will take this as the definition of d-separation and d-CG in the following.
\item The \emph{coarsest} $\sigma$-CG structure of a mixed graph $G$ is given by $\sigma(v):=\Sc^G(v) := \Anc^G(v) \cap \Desc^G(v)$ w.r.t.\ the underlying directed graph.
Note that the definition of strongly connected component totally ignores the bi- and undirected edges of the $\sigma$-CG. 
\item In any $\sigma$-CG we will always have that $\sigma$-separation implies d-separation, since every $Z$-d-open path is also $Z$-$\sigma$-open because $\{v\} \ins \sigma(v)$.
\item If a $\sigma$-CG $G$ is acyclic (implying $\Sc^G(v)=\{v\}$) then $\sigma$-separation coincides with d-separation.
\end{enumerate}
\end{Rem}

We now want to ``hide'' or marginalise out the latent nodes $u \in U$ from the graph of any mSCM and represent their induced dependence structure with bidirected edges.

\begin{Def}[Induced $\sigma$-CG of a mSCM]
\label{ind-s-CG}
Let $M=(G^+,\Xcal,\Pr,g)$ be a \hyle{mSCM}{mSCM} with $G^+=(U \dot \cup V, E^+)$.
The \emph{induced $\sigma$-CG $G$ of $M$}, also referred to as the \emph{causal graph} $G$ of $M$ is defined as follows:
	 \begin{enumerate}
	 \setlength\itemsep{0em}
	   \item The nodes of $G$ are all $v \in V$, i.e.\ all observed nodes of $G^+$.
		 \item $G$ contains all the directed edges of $G^+$ whose endnodes are both in $V$, i.e.\ observed.  
		 \item $G$ contains the bidirected edge $v \huh w$ with $v,w \in V$ if and only if $v \neq w$ and there exists a $u \in U$ with $v,w \in \Ch^{G^+}(u)$, i.e.\ $v$ and $w$ have a common latent confounder.
		 \item $G$ contains no undirected edges.
		 \item We put $\sigma(v):=\Sc^{G}(v) = \Anc^G(v) \cap \Desc^G(v)$.
	 \end{enumerate}
	\end{Def}
	
\begin{Rem}	
Caution must be applied when going from $G^+$ to $G$: It is possible that three observed nodes $v_1,v_2,v_3$ have one joint latent common cause $u_1$, which can be read off $G^+$. This information will get lost when going from $G^+$ to $G$, as we will represent this with three bidirected edges. $G$ will nonetheless capture the conditional independence relations (see Theorem~\ref{mSCM-gdGMP-thm}).
\end{Rem}

We now present the most important ingredient for our constraint-based causal discovery algorithm, namely a generalized directed global Markov property that relates the underlying causal graph ($\sigma$-CG) $G$  of any mSCM $M$ to the conditional independencies of the observed random variables $(X_v)_{v \in V}$ via a $\sigma$-separation criterion.

\begin{Thm}[$\sigma$-Separation Criterion, see Corollary~\ref{mSCM-gdGMP}]
\label{mSCM-gdGMP-thm}
The observed variables $(X_v)_{v \in V}$ of any \hyle{mSCM}{mSCM} $M$ 
satisfy the \emph{$\sigma$-separation criterion} 
w.r.t.\ the induced $\sigma$-CG $G$.
In other words, for all subsets $W,Y,Z \ins V$ we have the implication:
\[ W \Indep^\sigma_G Y \given Z \;\implies\; X_W \Indep_\Pr X_Y \given X_Z. \]
\end{Thm}

\begin{figure}[h]
\centering
\begin{tikzpicture}[scale=.7, transform shape]
\tikzstyle{every node} = [draw,shape=circle,color=blue]
\node (v1) at (0,0) {$v_1$};
\node (v2) at (1.5,-1.5) {$v_2$};
\node (v3) at (3,0) {$v_3$};
\node (v4) at (4.5,-1.5) {$v_4$};
\node (v5) at (3,-3) {$v_5$};
\node (v6) at (7.5,0) {$v_6$};
\node (v7) at (6.5,-1.5) {$v_7$};
\node (v8) at (5.5,-3) {$v_8$};
\draw[-{Latex[length=3mm, width=2mm]}, bend right, color=blue] (v7) to (v6);
\draw[-{Latex[length=3mm, width=2mm]}, bend right, color=blue] (v6) to (v7);
\draw[{Latex[length=3mm, width=2mm]}-{Latex[length=3mm, width=2mm]},red,out=245, in=205,looseness=1.5] (v1) to (v2);
\draw[{Latex[length=3mm, width=2mm]}-{Latex[length=3mm, width=2mm]}, red, bend left] (v4) to (v6);
\draw[{Latex[length=3mm, width=2mm]}-{Latex[length=3mm, width=2mm]},red,out=-25, in=-55,looseness=1.5] (v7) to (v6);
\foreach \from/\to in {v1/v2, v2/v3, v3/v4, v4/v5, v5/v2, v4/v8}
\draw[-{Latex[length=3mm, width=2mm]},color=blue] (\from) -- (\to);
\foreach \from/\to in {v7/v4} 
\draw[{Latex[length=3mm, width=2mm]}-{Latex[length=3mm, width=2mm]}, red] (\from) -- (\to);
\end{tikzpicture}
  \caption{The induced $\sigma$-connection graph (causal graph) of the modular structural causal model (mSCM) of figure \ref{fig:mSCM-graph}, with $\sigma$-equivalence classes $\{\{v_1\},\{v_2,v_3,v_4,v_5\},\{v_6,v_7\},\{v_8\}\}$. 
	}
 \label{fig:mSCM-sCG}
\end{figure}
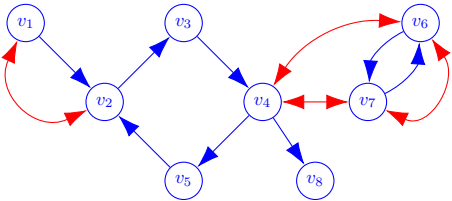

\begin{table}[h]
\caption{\label{tab:mSCM-indeps}d- and $\sigma$-separation in the $\sigma$-connection graph $G$ from figure 
 \ref{fig:mSCM-sCG}. $\sigma$-separation implies d-separation, but i.g.\ d-separation encodes more conditional independencies than
 $\sigma$-separation:}
\label{table:s-sep}
\begin{center}
\begin{tabular}{l|l}
 d-separation  & $\sigma$-separation \\
\hline 
$\{v_2\} \Indep^d_{G} \{v_4\} \given \{v_3,v_5\}$        &$\{v_2\} \nIndep^\sigma_{G} \{v_4\} \given \{v_3,v_5\}$ \\
$\{v_1\} \Indep^d_{G} \{v_6\} $        &$\{v_1\} \Indep^\sigma_{G} \{v_6\} $ \\
$\{v_1\} \Indep^d_{G} \{v_6\} \given \{v_3,v_5\}$        &$\{v_1\} \nIndep^\sigma_{G} \{v_6\} \given \{v_3,v_5\}$ \\
$\{v_1\} \nIndep^d_{G} \{v_8\}$        &$\{v_1\} \nIndep^\sigma_{G} \{v_8\}$ \\
$\{v_1\} \Indep^d_{G} \{v_8\} \given \{v_3,v_5\}$        &$\{v_1\} \nIndep^\sigma_{G} \{v_8\} \given \{v_3,v_5\}$ \\
$\{v_1\} \Indep^d_{G} \{v_8\} \given \{v_4\}$        &$\{v_1\} \Indep^\sigma_{G} \{v_8\} \given \{v_4\}$ \\
\hline
\end{tabular}
\end{center}
\end{table}

If we want to infer the causal graph ($\sigma$-CG) $G$ of a mSCM from data 
with help of conditional independence tests in practice we usually need to assume also the reverse implication of the $\sigma$-separation criterion from \ref{mSCM-gdGMP-thm} for the observational distribution $\Pr(X_V) $ and the relevant  
interventional distributions $\Pr(X_V|\doit(\xi_I))$ (see \ref{mSCM-marg-intv}). 
This will be called \emph{$\sigma$-faithfulness}.

\begin{Def}[$\sigma$-faithfulness]
\label{s-faithful}
 We will say that the tuple $(G,\Pr)$ is \hyle{s-faithful}{$\sigma$-faithful} if for every three subsets $W,Y,Z \ins V$ we have the equivalence:
\[ W \Indep^\sigma_G Y \given Z \;\iff\; X_W \Indep_\Pr X_Y \given X_Z. \]
\end{Def}


\begin{Rem}[Strong $\sigma$-completeness, cf.\ \cite{Meek95}]
We do believe that the \emph{generic} non-linear mSCM is $\sigma$-faithful to $G$, 
as the needed conditional dependence structure in all our simulated (sufficiently non-linear) cases was observed (cf.\ example \ref{the-example}).
But proving such a \emph{strong $\sigma$-completeness} result is difficult (and to our knowledge only done for multinomial and linear Gaussian DAG models, see \cite{Meek95}) and left for future research. Further note that the class of linear models, which even follow d-separation, would
i.g.\ not be considered $\sigma$-faithful. Since linear models are of measure zero in the bigger class of general mSCMs this would not contradict our conjecture.
\end{Rem}

\begin{Eg}
\label{the-example}
Consider a directed four-cycle, e.g.\ the subgraph $\{v_2,v_3,v_4,v_5\}$ from figure \ref{fig:mSCM-graph}, where all other observed nodes are assumed to be absent. 
Consider only the non-linear causal mechanisms given by ($i=3,4,5$):
\[
\begin{array}{rcrcl}
g_{\{v_2\}}(X_5,E_1) &:=&  \tanh\lp  0.9 \cdot X_5+0.5 \rp + E_1,\\
g_{\{v_i\}}(X_{i-1},E_i) &:=& \tanh\lp 0.9 \cdot X_{i-1}+0.5 \rp + E_i,\\
\end{array} \]
where $E_1,E_3,E_4,E_5$ are assumed to be independent. 
The equations $X_i = g_{\{v_i\}}(X_{i-1},E_i)$ and the one for $X_2$ will imply the conditional dependence
\[ X_2 \nIndep_{\Pr} X_4 \given (X_3,X_5), \]
which can be checked by computations and/or simulations. 
As one can read off table \ref{tab:mSCM-indeps}, d-separation fails to express the dependence, in contrast to $\sigma$-separation, which captures it correctly.
\end{Eg}

\subsection{MARGINALISATION AND CONDITIONING IN $\Sigma$-CONNECTION GRAPHS}

Inspired by \cite{HEJ14} we will define marginalisation and conditioning operations on $\sigma$-connection graphs ($\sigma$-CG) and prove the closedness of $\sigma$-separation (and thus its criterion) under these operations. These are key results to extend the algorithm of \cite{HEJ14} to the setting of mSCMs.

\begin{Def}[Marginalisation of a $\sigma$-CG]\label{def:marginalisation}
Let $G$ be a $\sigma$-CG with set of nodes $V$ and $w \in V$, $W:=\{w\}$.
We define the \emph{marginalised $\sigma$-CG} $G^W$ with set of nodes $V \sm W$ via the rules for $v_1,v_2 \in V \sm W$:\\
$v_1 \,{}_a \!\!\ouo\!\!{}_b\, v_2 \in G^W$ with arrow heads/tails $a$ and $b$ if and only if there exists:
\begin{enumerate}
\setlength\itemsep{0em}
 \item $v_1 \,{}_a\!\!\ouo\!\!{}_b\, v_2$ in $G$, or
 \item $v_1 \,{}_a\!\!\out w \,{}_q\!\!\ouo\!\!{}_b\, v_2$  in $G$, or
\item $v_1 \,{}_a\!\!\ouo\!\!{}_q\, w \,{}\!\!\tuo\!\!{}_b\, v_2$  in $G$, or
 \item $v_1 \,{}_a\!\!\ouh w \tut w \huo\!\!{}_b\, v_2$ in  $G$.
\end{enumerate}
Note that directed paths in $G$ have no colliders, so loops in $G$ map to loops in $G^W$ (if not empty). Thus we have the induced $\sigma$-CG structure $\sim_\sigma$ on $G^W$.
\end{Def}

\begin{Def}[Conditioning of a $\sigma$-CG]\label{def:conditioning}
Let $G$ be a $\sigma$-CG with set of nodes $V$ and $c \in V$, $C:=\{c\}$.
We define the \emph{conditioned $\sigma$-CG} $G_C$ with set of nodes $V \sm C$ via the rules for $v_1,v_2 \in V \sm C$:\\
$v_1 \,{}_a \!\!\ouo\!\!{}_b\, v_2 \in G_C$ if and only if there exists:
\begin{enumerate}
\setlength\itemsep{0em}
 \item $v_1 \,{}_a\!\!\ouo\!\!{}_b\, v_2$ in $G$, or
 \item $v_1 \,{}_a\!\!\ouh c \huo\!\!{}_b\, v_2$ in  $G$, or
 \item $v_1 \,{}_a\!\!\hut c \huo\!\!{}_b\, v_2$ in $G$, $\sigma(v_1)=\sigma(c)$, or
\item $v_1 \,{}_a\!\!\ouh c \tuh\!\!{}_b\, v_2$ in $G$, $\sigma(c)=\sigma(v_2)$, or
\item $v_1 \,{}_a\!\!\hut c \tuh\!\!{}_b\, v_2$ in $G$, $\sigma(v_1)=\sigma(c)=\sigma(v_2)$.
\end{enumerate}
Note that directed paths in $\sigma(v)$ in $G$ condition to directed paths, so loops in $\sigma(v)$ in $G$ map to loops in $G_C$ (if not empty). Thus we have a well-defined induced $\sigma$-CG structure $\sim_\sigma$ on $G_C$.
\end{Def}

The proofs of the following theorem, stating the closedness of $\sigma$-separation under marginalisation and conditioning, can be found in the supplementary material (Theorem~\ref{s-sep-marg-thm} and Theorem~\ref{s-sep-cond-thm}). See also figure \ref{fig:cCG-acyclic-cyclic}.

\begin{Thm}
\label{s-sep-marg-cond-thm}
Let $G$ be a $\sigma$-CG with set of nodes $V$ and $X,Y,Z \ins V$ any subsets.
For any nodes $w,c \in V\sm (X \cup Y \cup Z)$, $W:=\{w\}$, $C:=\{c\}$, we then have the equivalences:
\[\begin{array}{rcl} 
\displaystyle X \Indep^\sigma_{G^W} Y \given Z &\iff & \displaystyle X \Indep^\sigma_G Y \given Z, \\
\displaystyle X \Indep^\sigma_{G_C} Y \given Z &\iff & \displaystyle X \Indep^\sigma_G Y \given Z \cup C.
\end{array} \]
\end{Thm}

\begin{Cor}\label{cor:encoding}
Let $G$ be a $\sigma$-CG with set of nodes $V$ and $X,Y,Z\ins V$ pairwise disjoint subsets and $W:=V\sm (X \cup Y \cup Z)$.
Then we have the equivalence:
\[ X \Indep^\sigma_{G} Y \given Z \;\iff \; X \Indep^\sigma_{G_Z^W} Y,\]
where $G_Z^W$ is any $\sigma$-CG with set of nodes $X \cup Y$ obtained by marginalising out all the nodes from $W$ and conditioning on all the nodes from $Z$ in any order.
This means that if $X=\{x\}$ and $Y=\{y\}$ then $x$ and $y$ are $\sigma$-separated by $Z$ in $G$ if and only if $x$ and $y$ are not connected by any edge in the $\sigma$-CG $G_Z^W$.
\end{Cor}

It is also tempting to introduce an intervention operator directly on the level of $\sigma$-CGs.
However, since the interplay between conditioning and intervention is complicated (e.g.\ they do not commute i.g.) we do not investigate this further in this paper. The intervention operator on the level of mSCMs will be enough for our purposes as we assume no pre-interventional selection bias and then only encounter observational or post-interventional conditioning, which is covered by our framework.

\section{ALGORITHM}

In this section, we propose an algorithm for causal discovery that is based on the theory in the previous section.
Given that theory, our proposed algorithm is a straightforward modification of the algorithm by \cite{HEJ14}. The main idea is to formulate the causal discovery problem as an optimization problem that aims at finding the causal graph that best
matches the data at hand. This is done by encoding the rules for conditioning, marginalisation, and intervention (see below) on a 
$\sigma$-CG into Answer Set Programming (ASP), an expressive declarative programming language based on stable
model semantics that supports optimization \cite{Lifschitz08,Gelfond08}. 
The optimization problem can then be solved by employing an off-the-shelf ASP solver.

\subsection{CAUSAL DISCOVERY WITH $\Sigma$-CONNECTION GRAPHS}
Let $M=(G^+,\Xcal,\Pr,g)$ be a mSCM with $G^+=(U \dot \cup V, E^+)$ and $I \ins V$ a subset.
Consider a (stochastic) perfect intervention $\doit(\xi_I)$ that enforces $X_I = \xi_I$
for an independent random variable $\xi_I$ taking values in $\mathcal{X}_I$. 
Denote the (unique) induced distribution of the intervened mSCM $M_{\doit(\xi_I)}$ by $\Pr^{\doit(I)}$,
and the causal graph (i.e., induced $\sigma$-CG of the intervened mSCM on the observed variables) by 
$G_{\mathrm{do}(I)}=(G^+_{\mathrm{do}(I)})^U$. 

Under $\Pr^{\mathrm{do}(I)}$, the observed variables $(X_v)_{v \in V}$ satisfy the $\sigma$-separation criterion w.r.t.\ $G_{\mathrm{do}(I)}$ by Theorem~\ref{mSCM-gdGMP-thm}.
For the purpose of causal discovery, we will in addition assume $\sigma$-faithfulness (Definition~\ref{s-faithful}), i.e., that each conditional independence
between observed variables is due to a $\sigma$-separation in the causal graph. Taken together, and by applying
Corollary~\ref{cor:encoding}, we get for all subsets $W,Y,Z \ins V$ the equivalences:
\begin{equation}\label{eq:encoding_equivalence}\begin{split}
  X_W \Indep_{\Pr^{\mathrm{do}(I)}} X_Y \given X_Z & \;\iff\; W \Indep^\sigma_{G_{\mathrm{do}(I)}} Y \given Z \\
  & \;\iff\; W \Indep^\sigma_{(G_{\mathrm{do}(I)})_Z^{V\setminus W\cup Y \cup Z}} Y. 
\end{split}\end{equation}
If $W=\{w\}$ and $Y=\{y\}$ consist of a single node each, the latter
can be easily checked by testing whether $w$ is non-adjacent to $y$ in $(G_{\mathrm{do}(I)})_Z^{V\setminus W\cup Y \cup Z}$. 

\subsection{CAUSAL DISCOVERY AS AN OPTIMIZATION PROBLEM}
Following \cite{HEJ14}, we formulate causal discovery as an optimization problem where
a certain loss function is optimized over possible causal graphs. This loss function sums the
weights of all the inputs that are violated assuming a certain underlying causal graph.

The input for the algorithm is a list $S = \big((w_j,y_j,Z_j,I_j,\lambda_j)\big)_{j=1}^n$ of 
weighted conditional independence statements. Here, the weighted statement $(w_j,y_j,Z_j,I_j,\lambda_j)$
with $w_j,y_j \in V$, $Z_j, I_j \subseteq V$, and $\lambda_j \in \eRN := \RN \cup \{-\infty,+\infty\}$
encodes that $X_{w_j} \Indep_{\Pr^{\mathrm{do}(X_{I_j})}} X_{y_j} \given X_{Z_j}$ holds
with confidence $\lambda_j$, where a finite value of $\lambda_j$ gives a ``soft constraint''
and a value of $\lambda_j=\pm\infty$ imposes a ``hard constraint''. Positive weights 
encode that we have empirical support \emph{in favor} of the independence, whereas negative weights
encode empirical support \emph{against} the independence (in other words, in favor of
\emph{dependence}).

As in \cite{HEJ14}, we define a loss function that measures the amount of evidence
\emph{against} the hypothesis that the data was generated by an mSCM with causal graph 
$G$, 
by simply summing the absolute weights of the input statements that conflict with $G$ under
the $\sigma$-Markov and $\sigma$-faithfulness assumptions:
\begin{equation}\label{eq:loss}\begin{split}
  & \mathcal{L}(G, S) \\
  & := \sum_{(w_j,y_j,Z_j,I_j,\lambda_j)\in S} \lambda_j (\I_{\lambda_j > 0} - \I_{w_j \Indep^\sigma_{G_{\mathrm{do}(I_j)}} y_j \given Z_j})
\end{split}\end{equation}
where $\I$ is the indicator function. This loss function differs from the one used in \cite{HEJ14}
in that we use $\sigma$-separation instead of d-separation.
Causal discovery can now be formulated as the optimization problem:
\begin{equation}\label{eq:argmin}
  G^* = \argmin_{G\in\mathbb{G}(V)} \mathcal{L}_R(G, S)
\end{equation}
where $\mathbb{G}(V)$ denotes the set of all possible causal graphs with variables $V$.

The optimization problem (\ref{eq:argmin}) may have multiple optimal solutions, because the underlying causal graph may not be identifiable from the inputs. Nonetheless, some of the features of the causal graph (e.g., the presence or absence of a certain
directed edge) may still be identifiable. We employ the method proposed by \cite{MagliacaneClaassenMooij_NIPS_16} for scoring
the confidence that a certain feature $f$ is present by calculating the difference between the optimal losses under the 
additional hard constraints that the feature $f$ is present vs.\ that the feature $f$ is absent in $G$.

In our experiments, we will use the
weights proposed in \cite{MagliacaneClaassenMooij_NIPS_16}:
$\lambda_j = \log p_j - \log \alpha$, where $p_j$ is the p-value of a statistical test with independence
as null hypothesis, and $\alpha$ is a significance level (e.g., 1\%). This test is performed on the data 
measured in the context of the (stochastic) perfect intervention $I_j$. 
These weights have the desirable property that independences typically get a lower absolute weight
than strong dependences. For the conditional independence test, we use a standard partial correlation test after marginal rank-transformation of the data so as to obtain marginals with standard-normal distributions.


\subsection{FORMULATING THE OPTIMIZATION PROBLEM IN ASP}
In order to calculate the loss function (\ref{eq:loss}), we make use of Corollary~\ref{cor:encoding}
to reduce the $\sigma$-separation test to a simple non-adjacency test in a conditioned and
marginalised $\sigma$-CG, as in (\ref{eq:encoding_equivalence}).
We do this by encoding $\sigma$-CGs, Theorem~\ref{s-sep-marg-cond-thm} and the marginalisation
and conditioning operations on $\sigma$-CGs (Definitions~\ref{def:marginalisation} and \ref{def:conditioning}) in ASP.
The details of the encoding are provided in the Supplementary Material.\footnote{The full source code for the algorithm and to reproduce our experiments is available under an open source license from \mbox{\url{https://github.com/caus-am/sigmasep}.}}
The optimization problem in (\ref{eq:argmin}) can then be solved straightforwardly by running an 
off-the-shelf ASP solver with as input the encoding and the weighted independence statements.

A more precise statement of the following result is provided in the Supplementary Material. The proof is basically the same
as the one given in \cite{MagliacaneClaassenMooij_NIPS_16}.
\begin{Thm}
The algorithm for scoring features $f$ is sound for oracle inputs and asymptotically consistent under mild assumptions.
\end{Thm}

\section{EXPERIMENTS}
\subsection{CONSTRUCTING MSCMS AND SAMPLING FROM MSCMS}

To construct a modular structural causal model (mSCM) in practice we need to specify the compatible system of functions $(g_S)_{ S \in \Lcal(G)}$. The following Theorem is helpful (and a direct consequence of Banach's fixed point theorem).
\begin{Thm}\label{mSCM-Lipschitz}
Consider the functions $g_{\{v\}}$ for the trivial loops $\{v\} \in \Lcal(G)$, $v \in V$ and assume the following contractivity condition:
\begin{enumerate}
\setlength\itemsep{0em}
\item[] For every non-trivial loop $S \in \Lcal(G)$ and for every value $x_{\Pa^{G^+}(S) \sm S}$
the multi-dimensional function: 
\[ x_S' \mapsto \lp g_{\{v\}}(x_{S \cap \Pa^{G^+}(v)\sm \{v\}}', x_{ \Pa^{G^+}(v)\sm S} ) \rp_{v \in S}   \]
is a contraction, i.e.\ Lipschitz continuous with Lipschitz constant $L(x_{\Pa^{G^+}(S) \sm S})$ $<1$ w.r.t.\ a suitable norm $||\cdot||$.
\end{enumerate}
Then all the functions $g_S$ for the non-trivial loops $S \in \Lcal(G)$ exist, are unique and $g=(g_S)_{\substack{S \in \Lcal(G)}}$ forms a globally  compatible system. \\
More constructively, for every value $x_{\Pa^{G^+}(S) \sm S}$ and 
initialization $x_S^{(0)}$ the iteration scheme (using vector notations):
\[ x_S^{(t+1)} :=  (g_{\{v\}})_{v \in S}(x_S^{(t)},x_{\Pa^{G^+}(S) \sm S})  \]
converges to a unique limit vector $x_S$ (for $t \to \infty$ and independent of $x_S^{(0)}$). $g_S$ is then given by putting: 
\[g_S(x_{\Pa^{G^+}(S) \sm S}):= x_S.\]
\end{Thm}


This provides us with a method for constructing very general non-linear mSCMs (e.g.\ neural networks, see Section~\ref{nn-mSCM} in Supplementary Material) and to sample from them: by sampling $x_U$ from the external distribution and then apply the above iteration scheme until convergence for all loops, yielding the limit $x_V$ as one data point.



\subsection{RESULTS ON SYNTHETIC DATA}

\begin{figure}[h]
  \includegraphics[scale=0.5]{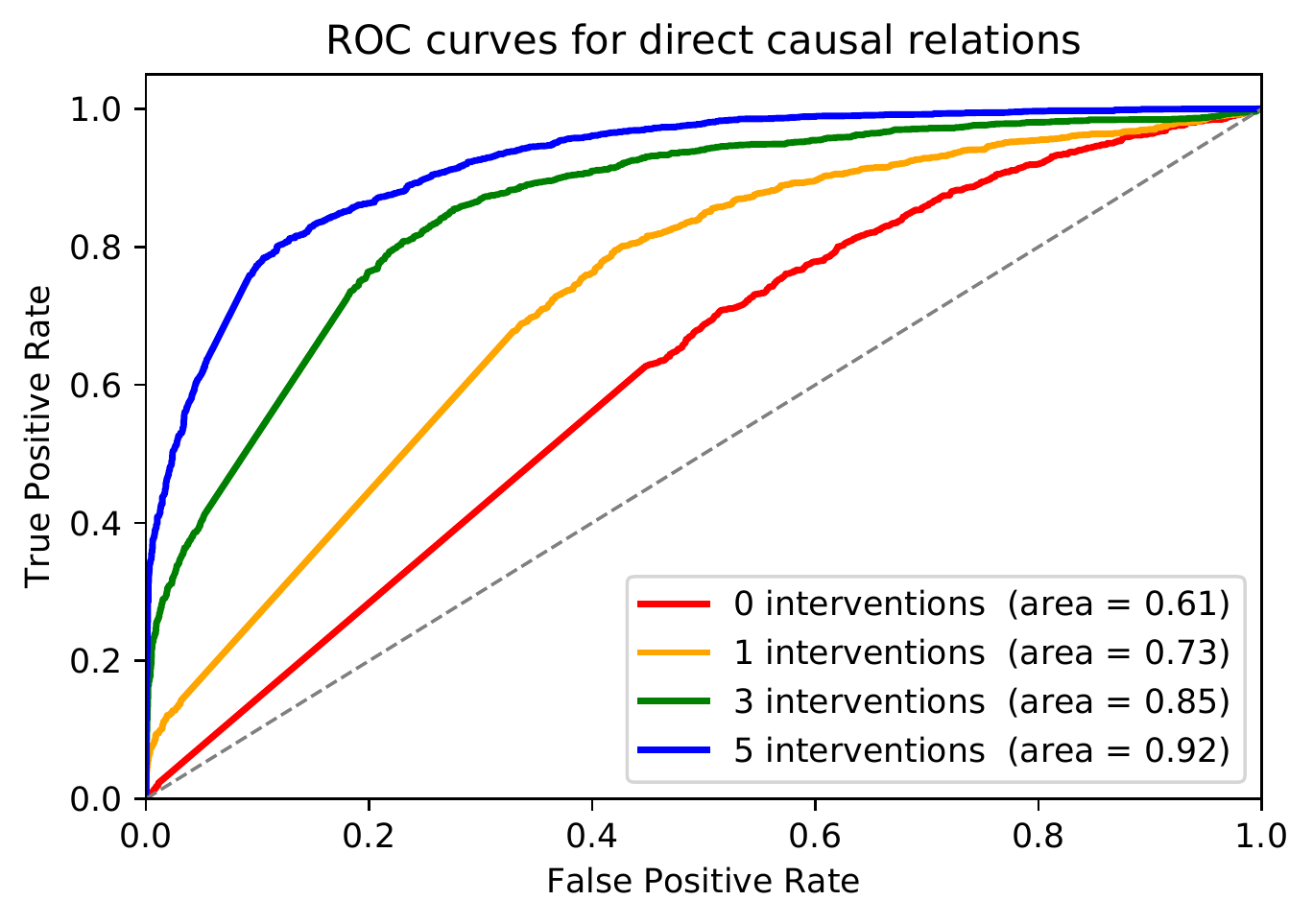}
	\caption{The ROC curves for identifying directed edges. See also figures~\ref{fig:roc-pr-ssep} and \ref{fig:roc-pr-encoding} in the Supplementary Material.}
	 \label{fig:roc}
\end{figure}

In our experiments we will---due to computational restrictions---only allow for $d=5$ observed nodes and $k=2$ additional latent confounders. We sample edges independently with a probability of $p=0.3$.
We model the non-linear function $g_{\{v\}}$ as a neural network with $\tanh$ activation, bias terms that have a normal distribution with mean $-0.5$ and standard deviation $0.2$, and weights sampled uniformly from the L1-unit ball to satisfy the contraction condition of Theorem~\ref{mSCM-Lipschitz} (see also Supplementary Material, Section~\ref{nn-mSCM}). 
We simulate 0--5 single-variable interventions with random (unique) targets.
For each intervened model we sample from standard-normal noise terms and compute the observations. To also detect weak dependencies in cyclic models we allow for $n=10^4$ samples in each such model for each allowed intervention. We then run all possible conditional independence tests between every pair of single nodes and calculate their $p$-values. We used $\alpha=10^{-3}$ as the threshold between dependence and independence. For computational reasons we restrict to partial correlation tests of marginal Gaussian rank-transforms of the data. These tests are then fed into the ASP solver together with our encoding of the optimization problem (\ref{eq:argmin}). We query the ASP solver for the confidence for the absence or presence of each possible directed and bidirected edge. We simulate $300$ models and aggregate results, using the confidence scores to compute ROC- and PR-curves for features. Figure \ref{fig:roc} shows that, as expected, our algorithm recovers more directed edges of the underlying causal graph in the simulation setting as the number of single-variable interventions increases. More results (ROC- and PR-curves for directed edges and confounders for different numbers of single-variable interventions and for different encodings) are provided in the Supplementary Material. 

\section{CONCLUSION}


We introduced $\sigma$-connection graphs ($\sigma$-CG) as a generalization of the d-connection graphs (d-CG) of \cite{HEJ14} and extended the notion of $\sigma$-separation that was introduced in \cite{FM17} to $\sigma$-CGs. We showed how $\sigma$-CGs behave under marginalisation and conditioning. This provides a graphical representation of how conditional independencies of modular structural causal models (mSCMs) behave under these operations. We provided a sufficient condition that allows constructing mSCMs and sampling from them.
We extended the algorithm of \cite{HEJ14} to deal with the more generally applicable notion of $\sigma$-separation instead of d-separation, thereby obtaining the first algorithm for causal discovery that can deal with cycles, non-linearities, latent confounders and a combination of data sets corresponding to observational and different interventional settings.
We illustrated the effectiveness of the algorithm on simulated data.
In this work, we restricted attention to (stochastic) perfect (``surgical'') interventions, but a straightforward extension to deal with other types of interventions and to generalize the idea of randomized controlled trials can be obtained by applying the JCI framework \cite{Mooij++_1611.10351v3}. In future work we wish to improve our algorithm to also handle selection bias, become more scalable and apply it to real world data sets.

\subsubsection*{Acknowledgements}
This work was supported by the European Research Council (ERC) under the European Union's Horizon 2020 research and innovation programme (grant agreement 639466).

%
\newpage
\bibliographystyle{plain}
\small

\newpage
\appendix
\normalsize
{\Large{\bf SUPPLEMENTARY MATERIAL}}


\section{$\Sigma$-CG UNDER MARGINALISATION AND CONDITIONING}


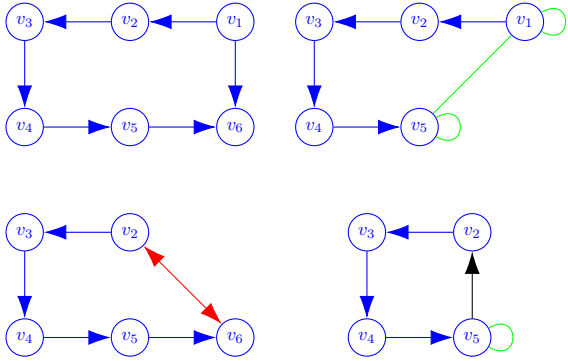
\begin{figure}[h]
${ }$\quad
\begin{tikzpicture}[scale=.7, transform shape]
  \tikzstyle{every node} = [draw,shape=circle,color=blue]
  \begin{scope}
    \node (v1) at (4,2) {$v_1$};
    \node (v2) at (2,2) {$v_2$};
    \node (v3) at (0,2) {$v_3$};
    \node (v4) at (0,0) {$v_4$};
    \node (v5) at (2,0) {$v_5$};
    \node (v6) at (4,0) {$v_6$};
    \foreach \from/\to in {v1/v2, v2/v3, v3/v4, v4/v5, v5/v6, v1/v6}
    \draw[-{Latex[length=3mm, width=2mm]},color=blue] (\from) -- (\to);
  \end{scope}
  \begin{scope}[xshift=5.5cm]
    \node (v1) at (4,2) {$v_1$};
    \node (v2) at (2,2) {$v_2$};
    \node (v3) at (0,2) {$v_3$};
    \node (v4) at (0,0) {$v_4$};
    \node (v5) at (2,0) {$v_5$};
    \draw[-, green] (v1) to (v5);
    \draw[-, green] (v1) edge[loop,in=-30,out=30,looseness=5] (v1);
    \draw[-, green] (v5) edge[loop,in=-30,out=30,looseness=5] (v5);
    \foreach \from/\to in {v1/v2, v2/v3, v3/v4, v4/v5}
    \draw[-{Latex[length=3mm, width=2mm]},color=blue] (\from) -- (\to);
  \end{scope}
  \begin{scope}[yshift=-4cm]
    \node (v2) at (2,2) {$v_2$};
    \node (v3) at (0,2) {$v_3$};
    \node (v4) at (0,0) {$v_4$};
    \node (v5) at (2,0) {$v_5$};
    \node (v6) at (4,0) {$v_6$};
    \draw[{Latex[length=3mm, width=2mm]}-{Latex[length=3mm, width=2mm]}, red] (v2) to (v6);
    \foreach \from/\to in { v2/v3, v3/v4, v4/v5, v5/v6}
    \draw[-{Latex[length=3mm, width=2mm]},color=blue] (\from) -- (\to);
  \end{scope}
  \begin{scope}[xshift=6.5cm,yshift=-4cm]
    \node (v2) at (2,2) {$v_2$};
    \node (v3) at (0,2) {$v_3$};
    \node (v4) at (0,0) {$v_4$};
    \node (v5) at (2,0) {$v_5$};
    \draw[-{Latex[length=3mm, width=2mm]}, color=black] (v5) to (v2);
    \draw[-, green] (v5) edge[loop,in=-30,out=30,looseness=5] (v5);
    \foreach \from/\to in {v2/v3, v3/v4, v4/v5}
    \draw[-{Latex[length=3mm, width=2mm]},color=blue] (\from) -- (\to);
  \end{scope}
\end{tikzpicture}
  \caption{A directed \emph{acyclic} graph $G$ (top left) as the $\sigma$-connection graph ($\sigma$-CG) with the $\sigma$-equivalence classes
	$\{ \{v_1\},\cdots,\{v_6\}\}$. We have the $\sigma$-separation:
	 $\{v_3\} \Indep^\sigma_G \{v_5\} \given \{v_2, v_4, v_6\}$.
		The combination of marginalizing out $v_1$ and conditioning on $v_6$ introduces cycles.
		Without keeping track of aboves $\sigma$-equivalence classes we would not get the corresponding $\sigma$-separation:
		$\{v_3\} \Indep^\sigma_{G^{\{v_1\}}_{\{v_6\}}} \{v_5\} \given \{v_2, v_4\}$ in the bottom right $\sigma$-CG.
		}
 \label{fig:cCG-acyclic-cyclic}
\end{figure}

\begin{Thm}[$\sigma$-Separation under Marginalisation]
\label{s-sep-marg-thm}
Let $G$ be a $\sigma$-CG with set of nodes $V$ and $W,X,Y,Z \ins V$ subsets with $W=\{w\}$  and $w \notin X\cup Y \cup Z$.
Then we have the equivalence:
\[ X \Indep^\sigma_G Y \given Z \; \iff \; 
X \Indep^\sigma_{G^W} Y \given Z.
\]
\begin{proof}
If $\pi=x\cdots y$ is a $Z$-$\sigma$-open path in $G$ then every occurrence of $w$ in $\pi$ is as a non-collider. 
If we have $\cdots v \tuh w \tuo \cdots$ in $\pi$ and $v \notin Z$ then marginalising out $w$ keeps $\pi^W$ $Z$-$\sigma$-open in $G^W$.
If $v \in Z$ then $v \in \sigma(w)$ by the $Z$-$\sigma$-openess.
Since $\sigma(w) \in \Lcal(G)$ is a loop we find elements $v_i \in \sigma(w)$, $r \ge 0$, $i=1,\dots r$, and a path
\[  \cdots v \hut v_1 \hut \cdots \hut v_r \hut w \tuo \cdots. \]
We do the same replacement on the right hand side of $w$ if necessary. 
Then marginalising this path w.r.t.\ $W$ gives a $Z$-$\sigma$-open path in $G^W$. This shows:
\[ X \Indep^\sigma_{G^W} Y \given Z \; \implies \; X \Indep^\sigma_{G} Y \given Z.\]
Now let $x \cdots y$ be a $Z$-$\sigma$-open path in $G^W$. Then every edge lifts to a subpath in $G$ where $w$ only occurs as a non-collider.
If a path $\cdots v \hut z \cdots$ in $G^W$ with $z \in Z \cap \sigma(v)$ in $G$ comes from $\cdots v \hut w \hut z \cdots$ or $v \huh w \tut w \hut z$ then we can, again since $\sigma(v) \in \Lcal(G)$ is a loop, find nodes $v_i \in \sigma(v)$, $r \ge 0$, $i=1,\dots r$, and a path in $G$ of the form:
\[ \cdots v \tuh v_1 \tuh \cdots \tuh v_r \tuh z \cdots \]
which is in any case $Z$-$\sigma$-open in $G$ (whether $v$ or $z$ are colliders or not).
So we can construct a $Z$-$\sigma$-open path in $G$ and we get:
\[ X \Indep^\sigma_{G} Y \given Z \; \implies \; X \Indep^\sigma_{G^W} Y \given Z.\]
\end{proof}
\end{Thm}

\begin{Thm}[$\sigma$-Separation under Conditioning]
\label{s-sep-cond-thm}
Let $G$ be a $\sigma$-CG with set of nodes $V$ and $C,X,Y,Z \ins V$ subsets with $C=\{c\}$  and $c \notin X\cup Y \cup Z$.
Then we have the equivalence:
\[ X \Indep^\sigma_G Y \given Z \cup C \; \iff \; 
X \Indep^\sigma_{G_C} Y \given Z.
\]
\begin{proof}
Let $C=\{c\}$.
Let $\pi=x \cdots y$ be a $(Z \cup C)$-$\sigma$-open path in $G$ with a minimal number of arrowheads pointing to nodes in $C$. 
Then at $c$ (if $c$ occurs) there are no undirected edges. So we have the cases:
\begin{enumerate}
\item fork: $\cdots v_1 \hut c \tuh v_2 \cdots$ in $G$ with $\sigma(v_1)=\sigma(c)=\sigma(v_2)$. Then $\cdots v_1 \huh v_2 \cdots$ is in $G_C$ with $\sigma(v_1)=\sigma(v_2)=\sigma(c)$. So the triple situation for $v_1$, $v_2$ stays the same in $G_C$.
\item right chain: $\cdots v_1 \begin{array}{c}\tuh\\[-10pt]\huh\end{array} c \tuh v_2 \cdots $ in $G$ with $\sigma(c)=\sigma(v_2)$. Then 
$\cdots v_1 \begin{array}{c}\tuh\\[-10pt]\huh\end{array} v_2 \dots$ is in $G_C$ with $\sigma(v_2)=\sigma(c)$. So the triple situation for $v_1$, $v_2$ stays the same in $G_C$.
\item left chain: similar to right chain.
\item collider: $\cdots v_1 \begin{array}{c}\huh\end{array} c \begin{array}{c}\huh\end{array}v_2 \cdots $ in $G$. Then $\cdots v_1 \huh v_2 \cdots$ is in $G_C$. So the triple situation for $v_1$, $v_2$ stays the same in $G_C$.
\item collider: $\cdots v_1 \begin{array}{c}\tuh\end{array} c \begin{array}{c}\hut\end{array}v_2 \cdots $ in $G$ with $v_1,v_2 \notin Z$.
Then $\cdots v_1 \tut v_2 \cdots $ is in $G_C$ with $v_1$, $v_2$ non-collider. So it is $Z$-open at $v_1$, $v_2$.
\item collider: $\cdots \dars v_1 \begin{array}{c}\tuh\end{array} c \begin{array}{c}\hut\\[-10pt]\huh\end{array}v_2 \cdots $ in $G$ with $v_1 \in Z \cap \sigma(c)$. Since $\sigma(c) \in \Lcal(G)$ is a loop there is a path in $G$ with $w_i \in \sigma(c)$, $r \ge 0$, $i=1,\dots,r$, of the form:
\[\cdots \dars v_1 \hut w_1 \hut[dashed] w_r \hut c \begin{array}{c}\hut\\[-10pt]\huh\end{array}v_2 \cdots,\] 
which then is $(Z \cup C)$-$\sigma$-open. So the path
\[\cdots \dars v_1 \hut w_1 \hut \cdots w_r  \begin{array}{c}\hut\\[-10pt]\huh\end{array}v_2 \cdots \]
 is then $Z$-$\sigma$-open in $G_C$.
\item collider: $\cdots v_1 \begin{array}{c}\tuh\end{array} c \begin{array}{c}\huh\end{array}v_2 \cdots $ in $G$ with $v_1 \notin Z$.
Then $\cdots v_1 \tuh v_2 \cdots $ is in $G_C$ and $Z$-open.
\item collider: as before with $v_1$ and $v_2$ swapped. Same arguments. 
\end{enumerate}
These cover all cases and we have shown:
\[ X \Indep_{G_C}^\sigma Y \given Z \; \implies\; X \Indep_G^\sigma Y \given Z \cup C.  \]
Now let $x \cdots y$ be a $Z$-open path in $G_C$.
Then the rules for conditioning lift every edge in $G_C$ to an edge or triple in $G$, where the triple situation for $c$ is $C$-$\sigma$-open and where the triple situation for the endnodes stays the same. So it is clearly $(Z \cup C)$-open in $G$. This shows:
\[ X \Indep_{G}^\sigma Y \given Z \cup C \; \implies\; X \Indep_{G_C}^\sigma Y \given Z.  \]
%
 %
%
\end{proof}
\end{Thm}

\section{THE $\Sigma$-SEPARATION CRITERION FOR MSCMS}

The trick to prove the $\sigma$-separation criterion is to transform the $\sigma$-connection graph $G$ of an mSCM, which has no undirected edges and can be seen as a directed mixed graph (DMG), into an acyclic directed mixed graph (ADMG) that encodes the same conditional independencies in terms of the well known d-separation. This also shows that every $\sigma$-separation-equivalence-class contains an acyclic graph (if one only looks at the observational distributions). Caution: the constructed ADMG is not well-behaved under marginalisation or interventions.
We will refer to the d-separation criterion as the \hyle{dGMP}{directed global Markov property (dGMP)} and to the $\sigma$-separation criterion as the 
\hyle{gdGMP}{generalized directed global Markov property (gdGMP)} in the following.

\begin{Lem}
\label{Lem-ADMG-dGMP}
Let $G=(V,E,H)$ be an acyclic directed mixed graph (ADMG) and $(X_v)_{v\in V}$ be random variables that satisfy the dGMP
w.r.t.\ $G$.
Let $E_w$ be a random variable independent of $(X_v)_{v\in V}$ and 
$X_w$ be another random variable, $w \notin V$, given by a functional relation:
\[ X_w=f\big((X_v)_{v \in P},E_w\big), \]
where $P \ins V$ is a subset of nodes.
Let $G'=(V',E',H')$ be the ADMG with set of nodes $V':=V \cup \{w\}$, set of edges $E':=E \cup \{ v \tuh w | v \in P\}$ and set of bidirected edges $H':=H$. 
Then $G \ins G'$ is an ancestral sub-ADMG and $(X_v)_{v\in V'}$ satisfies the dGMP
w.r.t.\ $G'$.
\begin{proof}
Since $w$ is a childless node in $G'$ clearly $G'$ is acyclic, $\Pa^{G'}(w)=P$ and $G \ins G'$ is an ancestral sub-ADMG.
So there exists a topological order $<$ for $G'$ such that $w$ is the last element.
Since for an ADMG the directed global Markov property (\hyle{dGMP}{dGMP}) is equivalent to the ordered local Markov property (\hyle{oLMP}{oLMP}) w.r.t.\ any topological order (see \cite{FM17,Richardson03}) we only need to check the local independence:
\[ \{w\} \Indep_\Pr A \sm \{w\} \given \partial_{A^\moral}(w)   \]
for every ancestral $A \ins G'$ with $w \in A$.
Since $\partial_{A^\moral}(w)=\Pa^{G'}(w)=P$ and $A \sm \{w\} \ins V$ the statement follows directly from the implication:
\begin{equation*}\begin{split}
  & E_w \Indep_{\Pr} (X_v)_{v \in V} \;\implies\\
& f\big((X_v)_{v \in P},E_w\big) \Indep_{\Pr} (X_v)_{v \in A \sm \{w\}} \given (X_v)_{v \in P}.  
\end{split}\end{equation*}
\end{proof}
\end{Lem}


\begin{Thm}
\label{rcsSEP-gdGMP}
Let $G=(V,E,H)$ be a directed mixed graph (DMG) and $\Scal(G)$ the set of its strongly connected components.
Assume that we have:
\begin{enumerate}
\item random variables $(X_v)_{v \in V}$,
\item random variables $(E_v)_{v \in V}$ that jointly satisfy  the \hyle{dGMP}{dGMP} w.r.t.\ the bidirected graph $(V,\emptyset,H)$, i.e.\
for every $W,Y \ins V$ we have the implication:
 \[ W \Indep^d_{(V,\emptyset,H)} Y \;\implies\; (E_v)_{v \in W} \Indep_\Pr (E_v)_{v \in Y}, \]
\item a tuple of functions $(g_S)_{S \in \Scal(G)}$ indexed by the strongly connected components $S$ of $G$,
\end{enumerate}
 such that we have the following equations for $S \in \Scal(G)$:
 \[  (X_v)_{v \in S} = g_S\big((X_w)_{w \in \Pa^G(S)\sm S},(E_w)_{w \in S}\big). \] 
Then $(X_v)_{v \in V}$ satisfy the \hyle{gdGMP}{general directed global Markov property (gdGMP)} w.r.t.\ the DMG $G$, i.e.\ for every three subsets $W,Y,Z \ins V$ we have the implication:
\[ W \Indep^\sigma_G Y \given Z \;\implies\; (X_v)_{v \in W} \Indep_\Pr (X_v)_{v \in Y} \given (X_v)_{v\in Z}. \]
\begin{proof}
By assumption we have that $(E_v)_{v \in V}$ satisfies the \hyle{dGMP}{dGMP} w.r.t.\ the ADMG $(V,\emptyset,H)$.
By lemma \ref{Lem-ADMG-dGMP} we can inductively add: 
 \[X_v=g_{S,v}\big((X_w)_{\Pa^G(S)\sm S},(E_w)_{w \in S}\big) \]
for $v \in V$ where $S=\Sc^G(v)$. We then finally get an ADMG $G'$ with nodes $(E_v)_{v\in V}$ and $(X_v)_{v \in V}$ that satisfy the \hyle{dGMP}{dGMP} w.r.t.\ this $G'$. This implies that for $W,Y,Z \ins V$ we have:
\[ W \Indep^d_{G'} Y \given Z \;\stackrel{}{\implies}\; (X_v)_{v \in W} \Indep_\Pr (X_v)_{v \in Y} \given (X_v)_{v\in Z}. \]
It is thus left to show that we also have the implication:
\[ W \Indep^\sigma_G Y \given Z \;\stackrel{}{\implies}\; W \Indep^d_{G'} Y \given Z. \]
For this it is enough to show that every $Z$-d-open path $\pi'$ from $W$ to $Y$ in $G'$ lifts to a $Z$-$\sigma$-open path $\pi$ from $W$ to $Y$ in $G$. The construction is straightforward. For details see \cite{FM17}.
\end{proof}
\end{Thm}

\begin{Cor}
\label{mSCM-gdGMP}
The observed variables $(X_v)_{v \in V}$ of any \hyle{mSCM}{mSCM} $M=(G^+,\Xcal,\Pr,g)$, $G^+=(U \dot \cup V, E^+)$, satisfy the
$\sigma$-separation criterion w.r.t.\ the induced $\sigma$-connection graph ($\sigma$-CG) $G$. 
\begin{proof}
For $v \in V$ we put $E_v:=(E_u)_{\substack{u \in U\\ v \in \Ch^{G^+}(u)}}$. The $(E_v)_{v \in V}$ then entail the conditional independence relations implied by d-separation of the bidirected graph $(V,\emptyset,H)$. Furthermore, for $S \in \Scal(G)$ we have equations:
\[X_S = g_{S}(X_{\Pa^{G}(S)\sm S},E_S). \]
The claim then directly follows from \ref{rcsSEP-gdGMP}.
\end{proof}
\end{Cor}

As a motivation for future work on selection bias we state the following direct corollary.

\begin{Cor}[mSCM with context]
\label{mSCM-gdGMP-cor}
Let $M=(G^+,\Xcal,\Pr,g)$ be a \hyle{mSCM}{mSCM} with $G^+=(U \dot \cup V, E^+)$ and $C \ins V$ a subset.
Let $G_C=(G^+)^U_C$ be the induced $\sigma$-CG of $M$ conditioned on $C$.
Then the observed variables $(X_v)_{v \in V\sm C}$ satisfy the $\sigma$-separation criterion 
 w.r.t.\ $G_C$ and w.r.t.\ the regular conditional probability distribution $\Pr^{|X_C=x_C}$ given $X_C=x_C$ (for $\Pr^{X_C}$-almost-all values $x_C\in \Xcal_C$):
For all subsets $W,Y,Z \ins V \sm C$ we have the implication:
\[ W \Indep^\sigma_{G_C} Y \given Z \;\implies\; X_W \Indep_{\Pr^{|X_C=x_C}} X_Y \given X_Z. \]
\begin{proof}
The lhs is equivalent to $W \Indep^\sigma_{G} Y \given Z \cup C$ (see Theorem~\ref{s-sep-marg-cond-thm}, Theorem~\ref{s-sep-cond-thm}, resp.) and this implies 
$X_W \Indep_{\Pr} X_Y \given X_Z, X_C$ (see Theorem~\ref{mSCM-gdGMP-thm}, Corollary~\ref{mSCM-gdGMP}, resp.), which implies the claim on the rhs (for $\Pr^{X_C}$-almost-all values $x_C \in \Xcal_C$).
\end{proof}
\end{Cor}

The last corollary can be used as a starting point for conditional independence constraint-based causal discovery in the presence of (unknown) \emph{selection bias} given by the unknown context $C$ and $x_C \in \Xcal_C$ (in addition to non-linear functional relations, cycles and latent confounders etc.).

\section{NEURAL NETWORKS AS MSCMS}
\label{nn-mSCM}


For constructing causal mechanisms we could use any parametric or non-parametric family of functions.
Since we want to stay as general as possible and also make use of the practical advantages of parametric models we 
represent/approximate the structural functions $g_{\{v\}}$, $v \in V$ by 
\emph{universal approximators}. A well known class of universal approximators are \emph{neural networks} (see 
e.g.\ \cite{Goodfellow-et-al-2016}). 
A neural network is a function that is constructed from several compositions of linear maps and a fixed one-dimensional \emph{activation} function $h$.
A sufficient condition to have the universal approximation property is if one assumes $h$ be continuous, non-polynomial and piecewise differentiable. 
A further advantage of neural networks is that the \emph{hidden units}  (given by composition of functions $z \mapsto h(w^Tz+b)$) can be interpreted as intermediate variables of an extended structural causal model.
This means that by modelling the hidden units of every $g_{\{v\}}$  explicitely as a node in an extended graph we can restrict---for the analysis purposes here---to this extended setting, where now the functions $g_{\{v\}}$ (the index $\{v\}$ refers to the trivial loop) are of the form:
\[\begin{array}{rcl}
&& g_{\{v\}}(x_{\Pa^{G^+}(v)\sm\{v\}}) \\
&= & h\lp \sum_{k \in \Pa^{G^+}(v)\sm\{v\}} A_{v,k} \cdot x_k +b_v \rp, 
\end{array}\]
with weights $A_{v,k}$ and biases $b_v$.

Further note that introducing or marginalizing intermediate variables will not change the outcome of the $\sigma$-separation criterion defined in Definition~\ref{s-sep-def}, Theorem~\ref{mSCM-gdGMP-thm}, and Theorem~\ref{s-sep-marg-cond-thm} (also see \cite{FM17}). So also this part is compatible with our theory. 

\begin{Thm}
The conditions for the contractiveness of the iterations scheme from subsection \ref{mSCM-Lipschitz} are satisfied if the following
 three points hold:
\begin{enumerate}
\setlength\itemsep{0em}
 \item $\sup_z|h'(z)| \le C $ with $0 < C < \infty$, and
 \item $A_{v,k} := 0$ for $k \notin \Pa^{G^+}(v) \sm \{v\}$, and
 \item $|| (A_{v,k})_{v,k \in S} || < \frac{1}{C}$ for every \emph{non-trivial} loop $S \ins G$, where $||\cdot||$ can be one of the matrix norms: $||\cdot||_p$, $p \ge 1$, or $||\cdot||_\infty$.
\end{enumerate}
In this case the functions $(g_{\{v\}})_{v \in V}$ will constitute a well-defined mSCM.
\end{Thm}
Note that we can put $C=1$ for popular activation functions $h(z)$ like $\tanh(z)$, $\ReLU(z)=\max(0,z)$, $\sigma(z)=\frac{1}{1+\exp(-z)}$, 
LeakyRelu, SoftPlus$(z)=\ln(1+e^z)$, etc..\\
Further note that by using one of these activation functions $h(z)$ and $||\cdot||=||\cdot||_\infty$ all the conditions are satisfied if we choose the $A_{v,k}$ such that for all $v \in V$:
 \[ \begin{array}{rcl} && \sum_{k \in \Pa^{G^+}(v)\sm \{v\}} |A_{v,k}| < 1 \\
& \text{ and } &   A_{v,k} := 0 \quad\text{ for }\quad k \notin \Pa^{G^+}(v) \sm \{v\}. \end{array} \]
Furthermore, we can then iterate the whole system for given error value $x_{U}$ and initialization $x_V^{(0)}$:
\[ \begin{array}{rcl}
 x_V^{(t+1)} &:=&  (g_{\{v\}})_{v \in V}(x_V^{(t)},x_{U})  \\
 &=& h\lp A_{V^+}\cdot ({x_V^{(t)} \atop x_{U}})+b_V \rp
\end{array}\]
and reach a unique fixed point $x_V$. This analyis also holds if we have the error variables outside of the activation function as additive noise.


\begin{proof}
\label{convergent}
For a non-trivial loop $S \ins G$ we want to show that for every value $x_{\Pa^{G^+}(S) \sm S}$ and initialization $x_S^{(0)}$ 
the iteration (using vector and matrix notations):
\[\begin{array}{rcl} && x_S^{(t+1)} := \\
&& h\lp A_{\Pa^{G^+}(S) \cup S} \cdot (x_S^{(t)},x_{\Pa^{G^+}(S) \sm S})^T + b_S\rp  \end{array}\]
converges to a unique point $x_S$ (for $t \to \infty$)
under the three stated assumptions in the text.\\
For applying Banach's fixed point theorem we need to show that for every value $x_{\Pa^{G^+}(S) \sm S}$ we have a bounded partial Jacobian ($S$ a \emph{non-trivial} loop):
\[ \sup _{x_S}||J_S(x_{\Pa^{G^+}(S) \cup S}) || \le L(x_{\Pa^{G^+}(S) \sm S}) < 1  \]
  where  $L(x_{\Pa^{G^+}(S) \sm S})$ is a constant smaller than $1$ and $||\cdot||$ is a suitable matrix norm. In our case we have: 
\[\begin{array}{rcl}
&&J_S(x_{\Pa^{G^+}(S) \cup S}) \\
&:=& \lp\frac{\partial g_{\{v\}}}{\partial x_k}\rp_{v,k \in S} (x_{\Pa^{G^+}(S) \cup S}) \\
&=& \nabla_{x_S} h\lp A_{\Pa^{G^+}(S) \cup S} \cdot \lp{x_S \atop x_{\Pa^{G^+}(S) \sm S}}\rp + b_S\rp \\
&=& \lp h'\lp\sum_{j \in \Pa^{G^+}(v)\sm\{v\}} A_{v,j} \cdot x_j + b_v\rp \right.\\
&& \left.\quad \cdot A_{v,k} \cdot \I_{k \in \Pa^{G^+}(v)\sm \{v\}} \rp_{v,k \in S} \\
&=& \diag(h') \cdot \lp A_S \odot \I_S \rp.
\end{array}
\]
Here $\diag(h')$ refers to the diagonal matrix with the corresponding values of $h'$  and $\I_S$ is the adjacency matrix as indicated on the line above. \\
If $ |h'(z)| \le C < \infty$ and $||\cdot||$ is either $||\cdot||_p$, $p \ge 1$, or $||\cdot||_\infty$ then
$||\diag(h')|| \le C$. If, furthermore, $|| A_S  \odot \I_S|| < \frac{1}{C}$ then we get:
\[\begin{array}{rcl}
|| J_S || &\le& || \diag(h')|| \cdot || A_S  \odot \I_S|| \\
&\le& C \cdot || A_S  \odot \I_S|| =: L \\
 & < & C \cdot \frac{1}{C}\\
 & = & 1.
\end{array}
\]
Note that we can represent $A  \odot \I$ in a single matrix $A$ if we put $A_{v,k}:=0$ whenever $k \notin \Pa^{G^+}(v)\sm \{v\}$.
From the above then follows that the map of the iteration scheme becomes contractive and the series thus converges to a unique fixed point $x_S$.
$g_S$ can then be defined via: 
\[g_S(x_{\Pa^{G^+}(S) \sm S}):= x_S.\]
The system $(g_S)_{S \in \Lcal(G)}$ is also compatible. Indeed, the convergence shows that the above element
$(x_{\Pa^{G^+}(S) \sm S}, x_S)$ simultaneously solves the system $x_v = g_{\{v\}}(x_{\Pa^{G^+}(v) \sm \{v\}})$, $v \in S$.
So for a loop $S' \ins S$ the corresponding components $(x_{\Pa^{G^+}(S') \sm S'}, x_{S'})$ simultaneously solves the system $x_v = g_{\{v\}}(x_{\Pa^{G^+}(v) \sm \{v\}})$, $v \in S'$. Since also the solution for the loop $S'$ is unique we get:
\[g_{S'}(x_{\Pa^{G^+}(S') \sm S'})= x_{S'},\]
which shows the compatibility.
The measurability of this map follows from a measurable choice theorem (see \cite{Bog07}) as explained in \cite{Bongers++_1611.06221v2}.
\end{proof}


If we want to uniformly sample weights for the parent nodes one can use the following:

\begin{Rem}[See \cite{BGMN05}]
To uniformly sample from the $d$-dimensional $L_p$-ball $B^d_p:=\{ x \in \R^d : ||x||_p \le 1 \}$
we can sample i.i.d.\ $y_1,\dots,y_d \sim p(t)=\frac{1}{2 \Gamma(1+\frac{1}{p})} e^{-|t|^p}$, $t \in \R$ and $z \sim p(s)=e^{-s}$, $s \ge 0$.
Then $x= \frac{(y_1,\dots,y_d)^T}{ \lp \sum_{j=1}^d |y_j|^p +z\rp^{1/p} }$ is uniformly sampled from $B^d_p$.
\end{Rem}

\section{MORE DETAILS ON THE ALGORITHM}


\subsection{SCORING FEATURES}

In order to score features, which can be defined as Boolean functions of the causal graph $G$, we define a modified loss function
\begin{equation}\label{eq:loss_feature}\begin{split}
  & \mathcal{L}(G, S, f) := \\
  & \sum_{(w_j,y_j,Z_j,I_j,\lambda_j)\in S} \lambda_j (\I_{\lambda_j > 0} - \I_{w_j \Indep^\sigma_{G_{\mathrm{do}(I_j)}} y_j \given Z_j} \I_{f(G)})
\end{split}\end{equation}
\cite{MagliacaneClaassenMooij_NIPS_16} proposed to score the confidence of a feature with
\begin{equation}\begin{split}\label{eq:confidence_feature}
  C(S,f)  := &\min_{G \in \mathbb{G}(V)} \mathcal{L}(G,S,\lnot f) \\
             & - \min_{G \in \mathbb{G}(V)} \mathcal{L}(G,S,f).
\end{split}\end{equation}
They showed that this scoring method is sound for oracle inputs.
\begin{Thm}\label{eq:soundness}
For any feature $f$, the confidence score $C(S,f)$ of (\ref{eq:confidence_feature}) is sound for oracle inputs with infinite weights. In other words,
$C(S,f)=\infty$ if $f$ is identifiable from the inputs,
$C(S,f)=-\infty$ if $\lnot f$ is identifiable from the inputs, and $C(S,f)=0$ if $f$ is unidentifiable from the inputs.
\end{Thm}

%

Furthermore, they showed that the scoring method is asymptotically consistent under a consistency condition on the statistical independence test. 
\begin{Thm}
Assume that the weights are asymptotically consistent, meaning that

\begin{equation}\label{eq:weightFreqConsistency}
\log p_N - \log \alpha_N \xto{P} \begin{cases}
  -\infty & H_1 \\
  +\infty & H_0,
\end{cases}
\end{equation}

%
as the number of samples $N \to \infty$, where the null hypothesis $H_0$ is independence and the alternative hypothesis
$H_1$ is dependence. Then for any feature $f$, the confidence score $C(S,f)$ of (\ref{eq:confidence_feature}) is asymptotically consistent, i.e., 
$C(S,f) \to \infty$ in probability if $f$ is identifiably true, $C(S,f) \to -\infty$ in probability
if $f$ is identifiably false, and $C(S,f)\to 0$ in probability otherwise.
\end{Thm}

By using the scoring method of \cite{MagliacaneClaassenMooij_NIPS_16} as explained above, our algorithm inherits these desirable properties.

\subsection{ENCODING IN ANSWER SET PROGRAMMING}
In order to test whether a causal graph $G$ entails a certain independence, we
create a computation graph of $\sigma$-connection graphs.  A computation graph
of $\sigma$-connection graphs is a DAG with $\sigma$-connection graphs as
nodes, and directed edges that correspond with the operations of conditioning
and marginalisation. The ``source node'' of an encoding DAG is an (intervened)
causal graph.  The ``sink'' nodes are $\sigma$-connection graphs that consist
of only two variables (because all other variables have been conditioned or
marginalised out) that can be reached from the source node by applying a
sequence of conditioning and marginalisation operations. Testing a
$\sigma$-separation statement in the intervened causal graph reduces to testing
for adjacency in the corresponding sink node.

Since interventions and conditioning do not commute, one has to take care to
employ these operations in the right ordering.  We define the computation graph
in such a way that intervention operations are performed first, followed by
marginalisations, and finally conditioning operations. At each stage, we always
remove the node with the highest possible label first, which means that our
computation graph is actually a computation tree. 

Below we provide the source code of the essential part of the algorithm, using the ASP syntax for \texttt{clingo 4}.
It is based upon the source code provided by \cite{HEJ14}. 
The differences to \cite{HEJ14}, i.e.\ of $\sigma$-separation vs. d-separation, are indicated with ``\emph{(sigma)}'' in the comments, i.e.\ at lines 100, 128, 138.
Note that the main difference between the encoding of d-separation and $\sigma$-separation is that in the non-collider case (see definition \ref{s-open-path}) we need to check in which strongly connected component $\sigma(v)$ the non-collider node lies in comparison to its adjacent nodes.
This boils down to checking ancestral relations. Since the $\sigma$-structure is inherited in a trivial fashion during the marginalisation and conditioning operations, it only needs to be found once (namely in the original $\sigma$-CG induced by the mSCM).

We used the state-of-the-art ASP solver \texttt{clingo 4} \cite{clingo} in our experiments to run the ASP program.

\onecolumn

\begin{lstlisting}
%%%%%%%%%% COMPUTATION GRAPH %%%%%%%%%%%%%%%%%%%%%%%%%%%%%%%

% preliminaries
node(0..nrnodes-1).
set(0..2**nrnodes-1).
ismember(M,Z) :- set(M), node(Z), M & (2**Z) != 0.

% first intervene, then marginalize, then condition;
%   always remove the node with the highest possible label first
intervene(0,Jsub,Z,J,0)   :- node(Z), set(J), set(Jsub), 
                             ismember(J,Z), not ismember(Jsub,Z), 
                             (Jsub + 2**Z) == J, 2**Z > Jsub.
marginalize(0,J,Msub,Z,M) :- node(Z), set(M), set(J), set(Msub),
                             ismember(M,Z), not ismember(Msub,Z),
                             (Msub + 2**Z) == M, 2**Z > Msub.
condition(Csub,Z,C,J,M)   :- node(Z), set(M), set(C), set(J), set(Csub),
                             (C & M) == 0, ismember(C,Z), not ismember(Csub,Z),
                             (Csub + 2**Z) == C, 2**Z > Csub.

% guess (generate all possible directed mixed graphs)
{ edge(X,Y) } :- node(X), node(Y), X != Y.
{ conf(X,Y) } :- node(X), node(Y), X < Y.

% source node in computation graph
th(X,Y,0,0,0) :- edge(X,Y).
hh(X,Y,0,0,0) :- conf(X,Y).

% ancestral relations after perfect (surgical) intervention on J
ancestor(X,Y,J) :- th(X,Y,0,J,0), X!=Y, node(X), node(Y), set(J).
ancestor(X,Y,J) :- ancestor(X,Z,J), ancestor(Z,Y,J), 
                   X!=Y, X!=Z, Y!=Z, 
                   node(X), node(Y), node(Z), set(J).


%%%%%%%%%% INTERVENTION %%%%%%%%%%%%%%%%%%%%%%%%%%%%%%%%%%%%

tt(X,Y,C,J,M) :- tt(X,Y,C,Jsub,M), 
                 X <= Y, 
                 not ismember(C,X), not ismember(C,Y),
                 not ismember(M,X), not ismember(M,Y),
                 node(X),node(Y),
                 intervene(C,Jsub,Z,J,M).

th(X,Y,C,J,M) :- th(X,Y,C,Jsub,M), 
                 X != Y,
                 not ismember(J,Y),
                 not ismember(C,X), not ismember(C,Y),
                 not ismember(M,X), not ismember(M,Y),
                 node(X),node(Y),
                 intervene(C,Jsub,Z,J,M).

hh(X,Y,C,J,M) :- hh(X,Y,C,Jsub,M), 
                 X < Y,
                 not ismember(J,Y), not ismember(J,X),
                 not ismember(C,X), not ismember(C,Y),
                 not ismember(M,X), not ismember(M,Y),
                 node(X),node(Y),
                 intervene(C,Jsub,Z,J,M).

%%%%%%%%%%%%%%%%%%%%%%%%%%%%%%%%%%%%%%%%%%%%%%%%%%%%%%%%%%


%%%%%%%%%% CONDITIONING %%%%%%%%%%%%%%%%%%%%%%%%%%%%%%%%%%

%% X---Y => X---Y
tt(X,Y,C,J,M) :- tt(X,Y,Csub,J,M), 
                 X <= Y, 
                 not ismember(C,X), not ismember(C,Y),
                 not ismember(M,X), not ismember(M,Y),
                 node(X),node(Y),
                 condition(Csub,Z,C,J,M).

%% X-->Z<--Y => X---Y
tt(X,Y,C,J,M) :- th(X,Z,Csub,J,M),th(Y,Z,Csub,J,M), 
                 X <= Y, 
                 not ismember(C,X), not ismember(C,Y),
                 not ismember(M,X), not ismember(M,Y),
                 node(X),node(Y),
                 condition(Csub,Z,C,J,M).

%%%%%%%%%%%%%%%%%%%%%%%%%%%%%%%%%%%%%%%%%%%%%%%%%%%%%%%%%%%%

%% X-->Y => X-->Y
th(X,Y,C,J,M) :- th(X,Y,Csub,J,M), 
                 X != Y,
                 not ismember(C,X), not ismember(C,Y),
                 not ismember(M,X), not ismember(M,Y),
                 node(X),node(Y),
                 condition(Csub,Z,C,J,M).

%% X-->Z<->Y => X-->Y
th(X,Y,C,J,M) :- th(X,Z,Csub,J,M),
                 { hh(Z,Y,Csub,J,M); hh(Y,Z,Csub,J,M) } >= 1,
                 X != Y, 
                 not ismember(C,X), not ismember(C,Y),
                 not ismember(M,X), not ismember(M,Y),
                 node(X),node(Y),
                 condition(Csub,Z,C,J,M).

%% X-->Z-->Y (anc of Z) => X-->Y (sigma)
th(X,Y,C,J,M):- th(X,Z,Csub,J,M), th(Z,Y,Csub,J,M),
                 ancestor(Y,Z,J),
                 X != Y,
                 not ismember(C,X), not ismember(C,Y),
                 not ismember(M,X), not ismember(M,Y),
                 node(X),node(Y),node(Z),
                 condition(Csub,Z,C,J,M).

%%%%%%%%%%%%%%%%%%%%%%%%%%%%%%%%%%%%%%%%%%%%%%%%%%%%%%%%%%%%

%% X<->Y => X<->Y
hh(X,Y,C,J,M) :- hh(X,Y,Csub,J,M), 
                 X < Y,
                 not ismember(C,X), not ismember(C,Y),
                 not ismember(M,X), not ismember(M,Y),
                 node(X),node(Y),
                 condition(Csub,Z,C,J,M).

%% X<->Z<->Y => X<->Y
hh(X,Y,C,J,M) :- { hh(Z,X,Csub,J,M); hh(X,Z,Csub,J,M) } >= 1,
                 { hh(Z,Y,Csub,J,M); hh(Y,Z,Csub,J,M) } >= 1,
                 X < Y, 
                 not ismember(C,X), not ismember(C,Y),
                 not ismember(M,X), not ismember(M,Y),
                 node(X),node(Y),
                 condition(Csub,Z,C,J,M).

%% X<->Z-->Y (anc of Z)  => X<->Y (sigma)
hh(X,Y,C,J,M) :- { hh(Z,X,Csub,J,M); hh(X,Z,Csub,J,M) } >= 1,
                 th(Z,Y,Csub,J,M),
                 ancestor(Y,Z,J),
                 X < Y,
                 not ismember(C,X), not ismember(C,Y),
                 not ismember(M,X), not ismember(M,Y),
                 node(X),node(Y),node(Z),
                 condition(Csub,Z,C,J,M).

%% (anc of Z) X<--Z-->Y (anc of Z) => X<->Y (sigma)
hh(X,Y,C,J,M) :- th(Z,X,Csub,J,M), th(Z,Y,Csub,J,M),
                 ancestor(X,Z,J),
                 ancestor(Y,Z,J),
                 X < Y,
                 not ismember(C,X), not ismember(C,Y),
                 not ismember(M,X), not ismember(M,Y),
                 node(X),node(Y),node(Z),
                 condition(Csub,Z,C,J,M).

%%%%%%%%%%%%%%%%%%%%%%%%%%%%%%%%%%%%%%%%%%%%%%%%%%%%%%%%%%%%


%%%%%%%%%% MARGINALIZATION %%%%%%%%%%%%%%%%%%%%%%%%%%%%%%%%%

%% X---Y => X---Y
tt(X,Y,C,J,M) :- tt(X,Y,C,J,Msub), 
                 X <= Y, 
                 not ismember(C,X), not ismember(C,Y),
                 not ismember(M,X), not ismember(M,Y),
                 node(X),node(Y),
                 marginalize(C,J,Msub,Z,M).

%% X-->Z---Y => X---Y
tt(X,Y,C,J,M) :- th(X,Z,C,J,Msub), 
                 { tt(Z,Y,C,J,Msub); tt(Y,Z,C,J,Msub) } >= 1,
                 X <= Y, 
                 not ismember(C,X), not ismember(C,Y),
                 not ismember(M,X), not ismember(M,Y),
                 node(X),node(Y),
                 marginalize(C,J,Msub,Z,M).

%% X---Z<--Y => X---Y
tt(X,Y,C,J,M) :- { tt(X,Z,C,J,Msub); tt(Z,X,C,J,Msub) } >= 1, 
                 th(Y,Z,C,J,Msub),
                 X <= Y, 
                 not ismember(C,X), not ismember(C,Y),
                 not ismember(M,X), not ismember(M,Y),
                 node(X),node(Y),
                 marginalize(C,J,Msub,Z,M).

%% X---Z---Y => X---Y
tt(X,Y,C,J,M) :- { tt(X,Z,C,J,Msub); tt(Z,X,C,J,Msub) } >= 1, 
                 { tt(Z,Y,C,J,Msub); tt(Y,Z,C,J,Msub) } >= 1,
                 X <= Y, 
                 not ismember(C,X), not ismember(C,Y),
                 not ismember(M,X), not ismember(M,Y),
                 node(X),node(Y),
                 marginalize(C,J,Msub,Z,M).

%% X-->Z---Z<--Y => X---Y
tt(X,Y,C,J,M) :- th(X,Z,C,J,Msub),th(Y,Z,C,J,Msub),tt(Z,Z,C,J,Msub), 
                 X <= Y, 
                 not ismember(C,X), not ismember(C,Y),
                 not ismember(M,X), not ismember(M,Y),
                 node(X),node(Y),
                 marginalize(C,J,Msub,Z,M).

%%%%%%%%%%%%%%%%%%%%%%%%%%%%%%%%%%%%%%%%%%%%%%%%%%%%%%%%%%%%

%% X-->Y => X-->Y
th(X,Y,C,J,M) :- th(X,Y,C,J,Msub), 
                 X != Y,
                 not ismember(C,X), not ismember(C,Y),
                 not ismember(M,X), not ismember(M,Y),
                 node(X),node(Y),
                 marginalize(C,J,Msub,Z,M).

%% X-->Z-->Y => X-->Y
th(X,Y,C,J,M) :- th(X,Z,C,J,Msub),th(Z,Y,C,J,Msub), 
                 X != Y,
                 not ismember(C,X), not ismember(C,Y),
                 not ismember(M,X), not ismember(M,Y),
                 node(X),node(Y),
                 marginalize(C,J,Msub,Z,M).

%% X---Z<->Y => X-->Y
th(X,Y,C,J,M) :- { tt(X,Z,C,J,Msub); tt(Z,X,C,J,Msub) } >= 1,
                 { hh(Z,Y,C,J,Msub); hh(Y,Z,C,J,Msub) } >= 1, 
                 X != Y,
                 not ismember(C,X), not ismember(C,Y),
                 not ismember(M,X), not ismember(M,Y),
                 node(X),node(Y),
                 marginalize(C,J,Msub,Z,M).

%% X---Z-->Y => X-->Y
th(X,Y,C,J,M) :- { tt(X,Z,C,J,Msub); tt(Z,X,C,J,Msub) } >= 1,
                 th(Z,Y,C,J,Msub), 
                 X != Y,
                 not ismember(C,X), not ismember(C,Y),
                 not ismember(M,X), not ismember(M,Y),
                 node(X),node(Y),
                 marginalize(C,J,Msub,Z,M).

%% X-->Z---Z<->Y => X-->Y
th(X,Y,C,J,M) :- th(X,Z,C,J,Msub), 
                 tt(Z,Z,C,J,Msub),
                 { hh(Z,Y,C,J,Msub); hh(Y,Z,C,J,Msub) } >= 1, 
                 X != Y,
                 not ismember(C,X), not ismember(C,Y),
                 not ismember(M,X), not ismember(M,Y),
                 node(X),node(Y),
                 marginalize(C,J,Msub,Z,M).

%%%%%%%%%%%%%%%%%%%%%%%%%%%%%%%%%%%%%%%%%%%%%%%%%%%%%%%%%%%%%

%% X<->Y => X<->Y
hh(X,Y,C,J,M) :- hh(X,Y,C,J,Msub),
                 X < Y,
                 not ismember(C,X), not ismember(C,Y),
                 not ismember(M,X), not ismember(M,Y),
                 node(X),node(Y),
                 marginalize(C,J,Msub,Z,M).

%% X<->Z-->Y => X<->Y
hh(X,Y,C,J,M) :- { hh(X,Z,C,J,Msub); hh(Z,X,C,J,Msub) } >= 1,
                 th(Z,Y,C,J,Msub),
                 X < Y,
                 not ismember(C,X), not ismember(C,Y),
                 not ismember(M,X), not ismember(M,Y),
                 node(X),node(Y),
                 marginalize(C,J,Msub,Z,M).

%% X<--Z-->Y => X<->Y
hh(X,Y,C,J,M) :- th(Z,X,C,J,Msub),
                 th(Z,Y,C,J,Msub),
                 X < Y,
                 not ismember(C,X), not ismember(C,Y),
                 not ismember(M,X), not ismember(M,Y),
                 node(X),node(Y),                                
                 marginalize(C,J,Msub,Z,M).

%% X<--Z<->Y => X<->Y
hh(X,Y,C,J,M) :- th(Z,X,C,J,Msub),
                 { hh(Y,Z,C,J,Msub); hh(Z,Y,C,J,Msub) } >= 1,
                 X < Y,
                 not ismember(C,X), not ismember(C,Y),
                 not ismember(M,X), not ismember(M,Y),
                 node(X),node(Y),
                 marginalize(C,J,Msub,Z,M).

%% X<->Z---Z<->Y => X<->Y
hh(X,Y,C,J,M) :- { hh(X,Z,C,J,Msub); hh(Z,X,C,J,Msub) } >= 1,
                 { hh(Y,Z,C,J,Msub); hh(Z,Y,C,J,Msub) } >= 1,
                 tt(Z,Z,C,J,Msub),
                 X < Y,
                 not ismember(C,X), not ismember(C,Y),
                 not ismember(M,X), not ismember(M,Y),
                 node(X),node(Y),
                 marginalize(C,J,Msub,Z,M).
                                
%%%%%%%%%%%%%%%%%%%%%%%%%%%%%%%%%%%%%%%%%%%%%%%%%%%%%%%%%%%%


%%%%%%%%% LOSS FUNCTION %%%%%%%%%%%%%%%%%%%%%%%%%%%%%%%%%%%%

fail(X,Y,C,J,M,W) :- th(X,Y,C,J,M), indep(X,Y,C,J,M,W), X<Y.
fail(X,Y,C,J,M,W) :- th(Y,X,C,J,M), indep(X,Y,C,J,M,W), X<Y.
fail(X,Y,C,J,M,W) :- hh(X,Y,C,J,M), indep(X,Y,C,J,M,W), X<Y.
fail(X,Y,C,J,M,W) :- tt(X,Y,C,J,M), indep(X,Y,C,J,M,W), X<Y.

fail(X,Y,C,J,M,W) :- not th(X,Y,C,J,M), 
                     not th(Y,X,C,J,M), 
                     not hh(X,Y,C,J,M), 
                     not tt(X,Y,C,J,M), 
                     dep(X,Y,C,J,M,W), X<Y.

%%%%%%%%%%%%%%%%%%%%%%%%%%%%%%%%%%%%%%%%%%%%%%%%%%%%%%%%%%%%


%%%%%%%%% OPTIMIZATION PROBLEM %%%%%%%%%%%%%%%%%%%%%%%%%%%%%

#minimize{W,X,Y,C,J,M:fail(X,Y,C,J,M,W) }.
\end{lstlisting}

\twocolumn
\section{EXPERIMENTAL RESULTS}
\label{supp-exp}

Here we provide additional visualisations of the results of our experiments, for which no space
was left in the main paper.

\begin{figure}[b]
  \includegraphics[scale=0.55]{ROC_edges_intv}\qquad
	\includegraphics[scale=0.55]{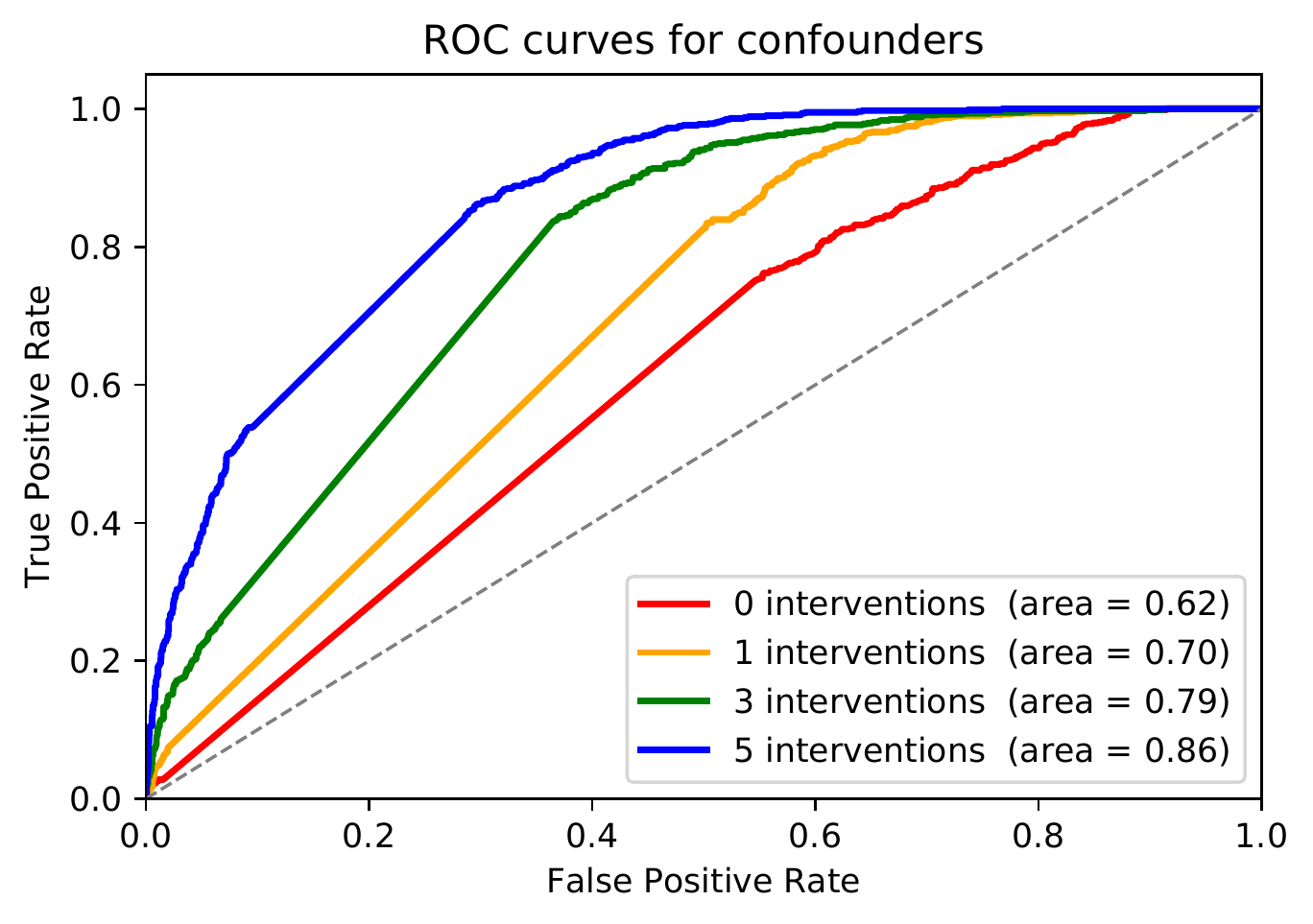}\\
  \\
	\includegraphics[scale=0.55]{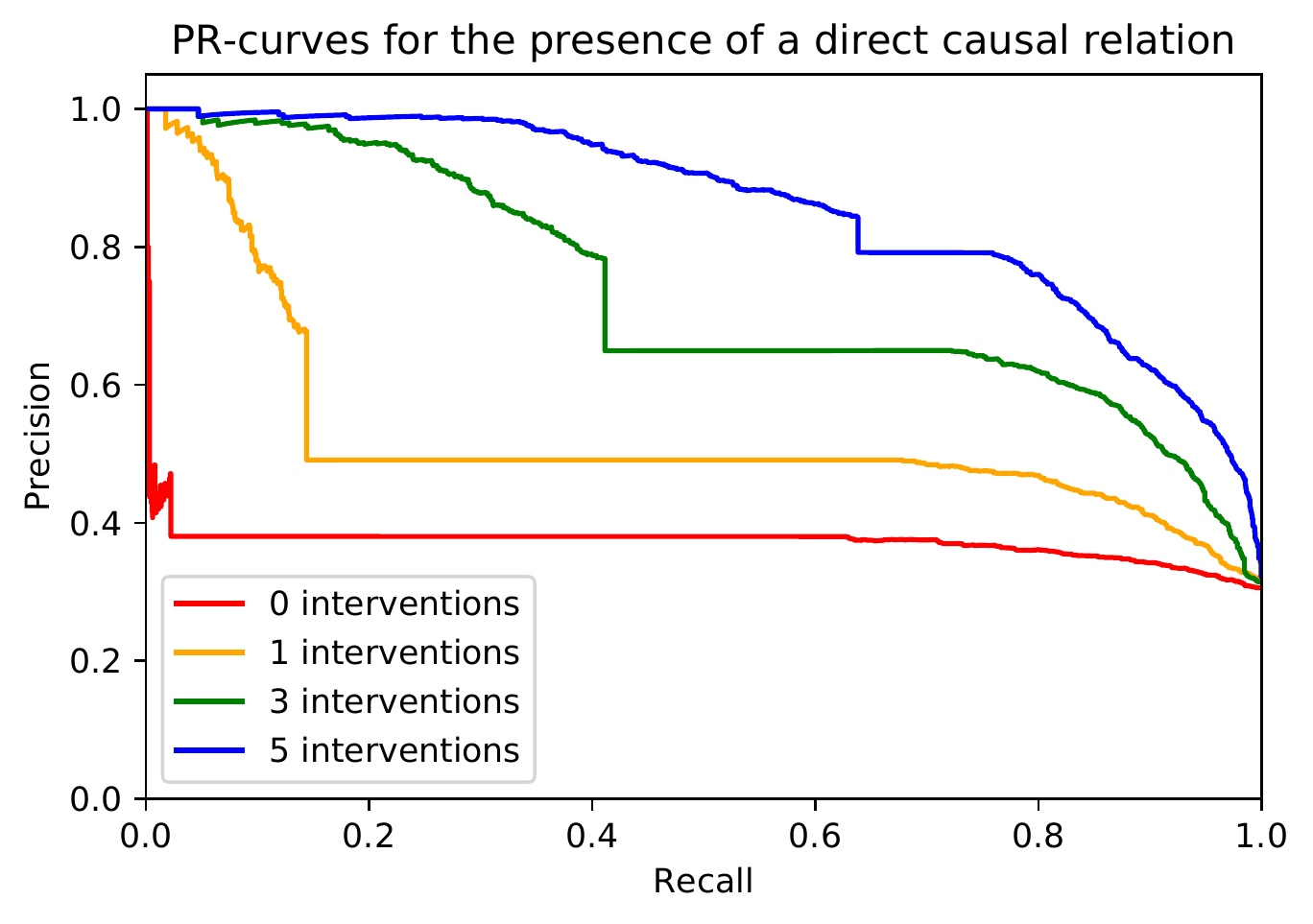}\qquad
	\includegraphics[scale=0.55]{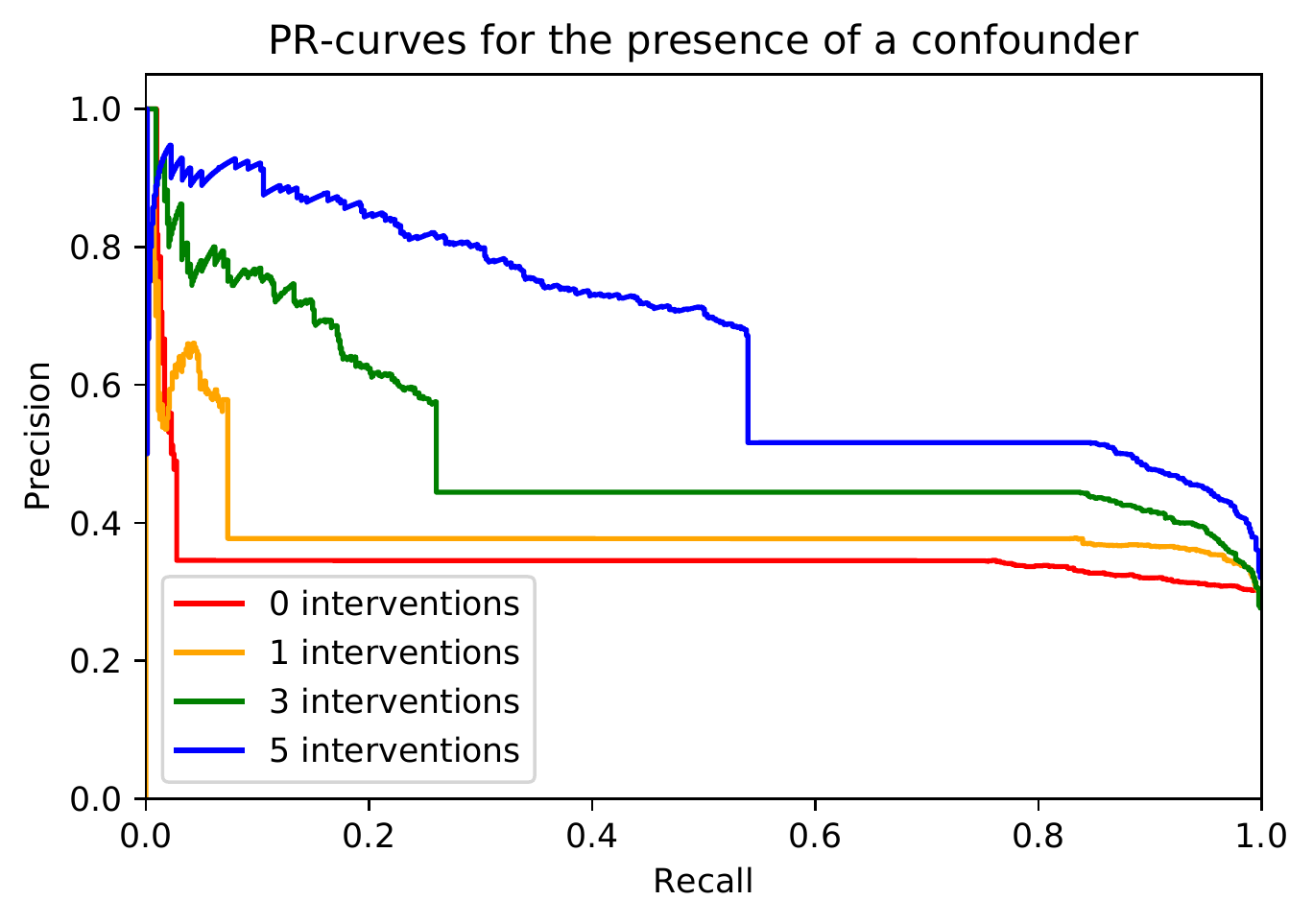}\\
  \\
	\includegraphics[scale=0.55]{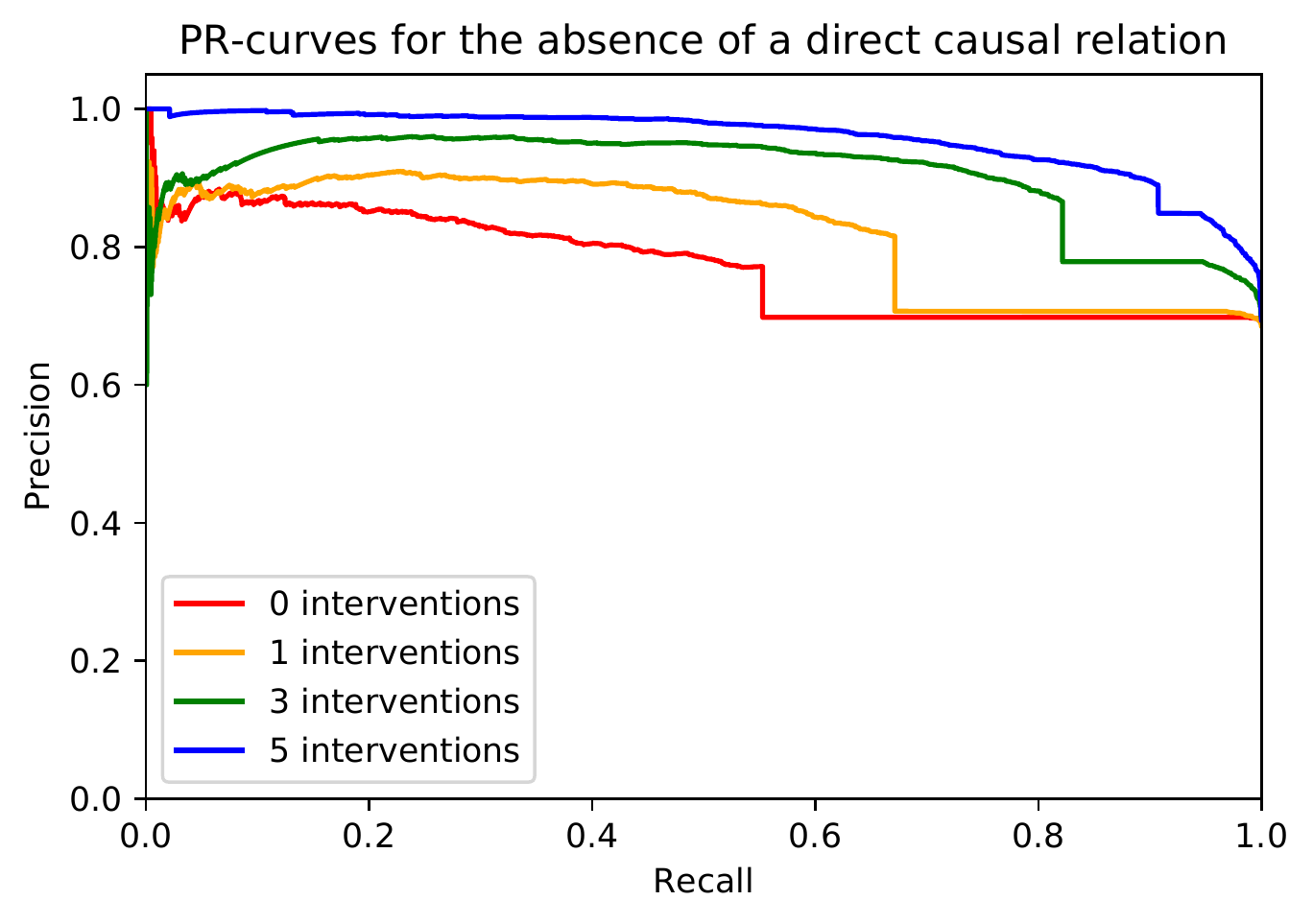}\qquad
	\includegraphics[scale=0.55]{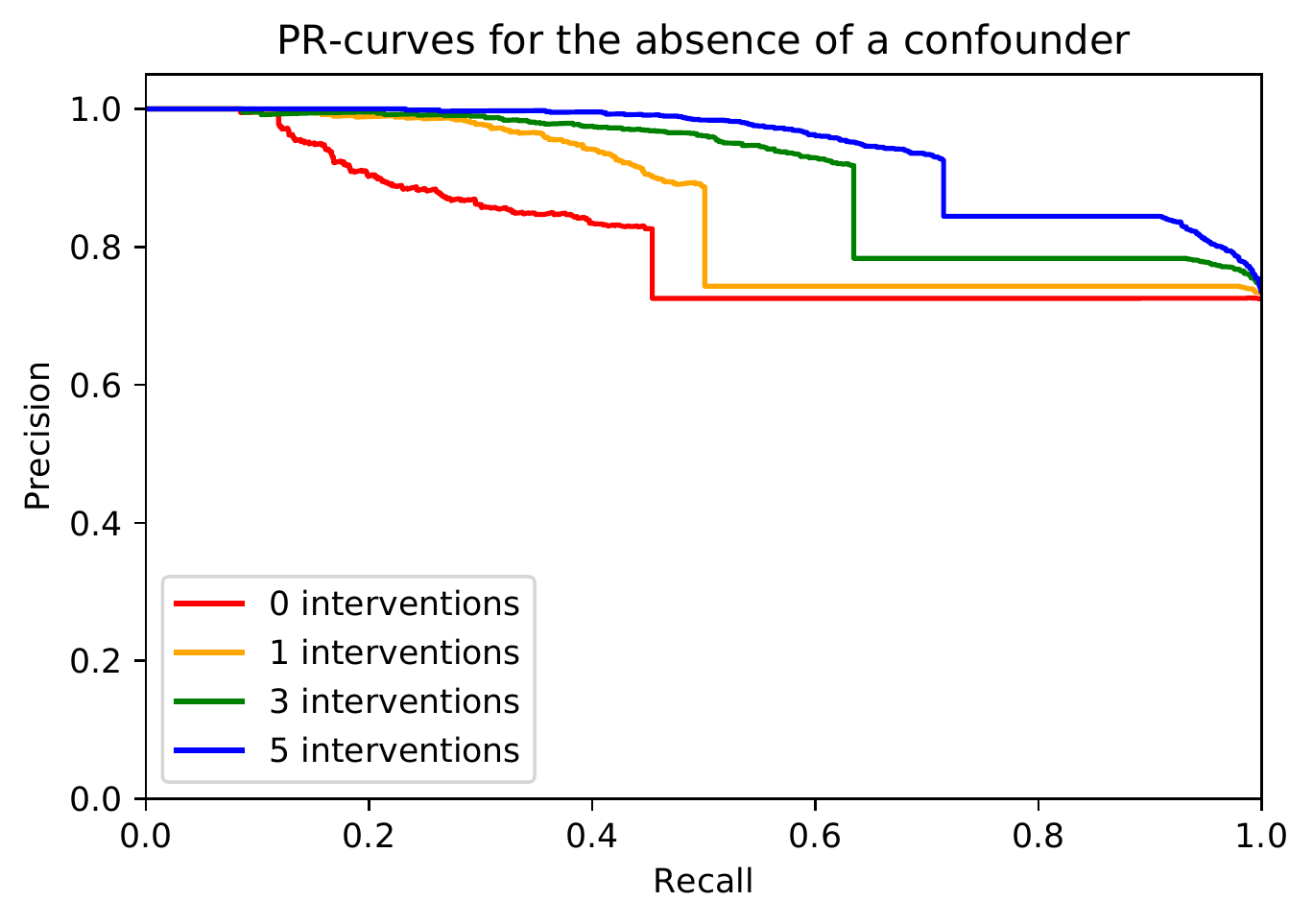}
  \caption{ROC curves (top) and PR curves (center, bottom) for directed edges (left) and confounders (right), for different numbers of single-variable interventions. All results shown here use $\sigma$-separation.}
	 \label{fig:roc-pr-ssep}
\end{figure}

Figure~\ref{fig:roc-pr-ssep} shows ROC curves and PR curves for detecting
directed edges (i.e., direct causal relations) and for detecting latent
confounders in the causal graph. Results are shown for the purely observational
setting (``0 interventions'') and for a combination of observational and
interventional data (``1--5 interventions'') where the targets of the
stochastic surgical interventions are single variables chosen randomly, without
replacement.  Clearly, making use of interventional data is beneficial for causal
discovery.

\begin{figure}[t]
  \includegraphics[scale=0.55]{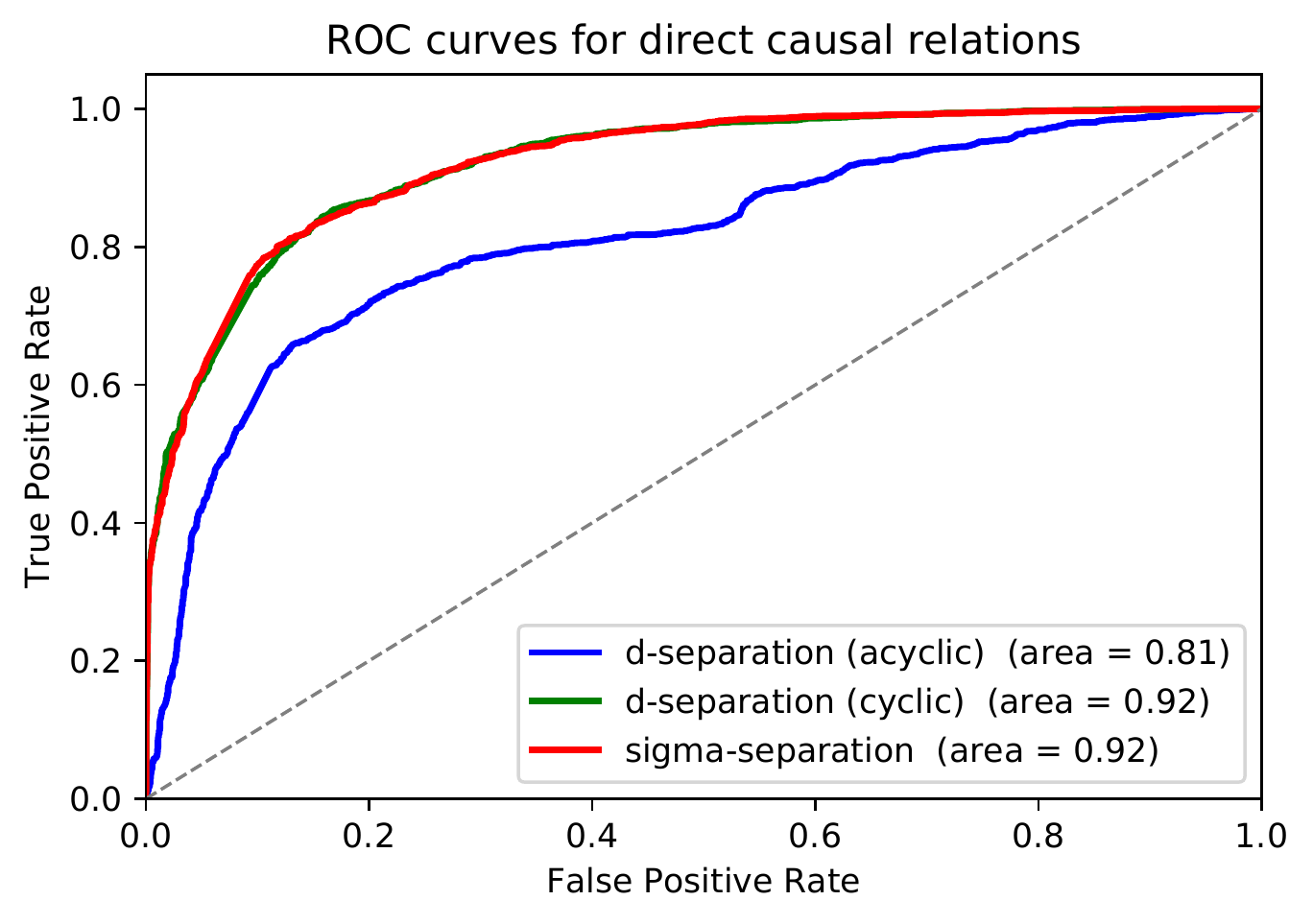}\qquad
	\includegraphics[scale=0.55]{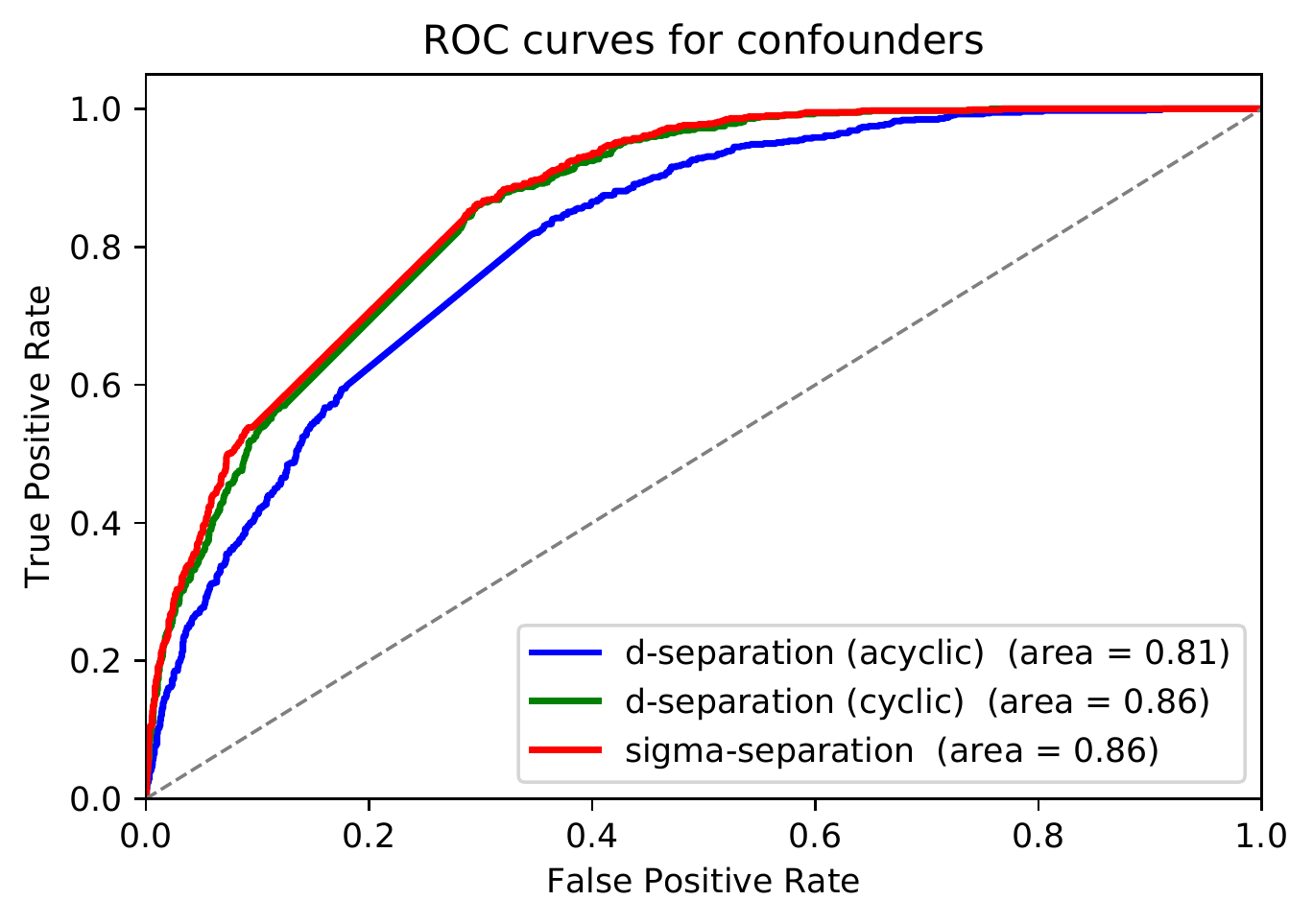}\\
  \\
	\includegraphics[scale=0.55]{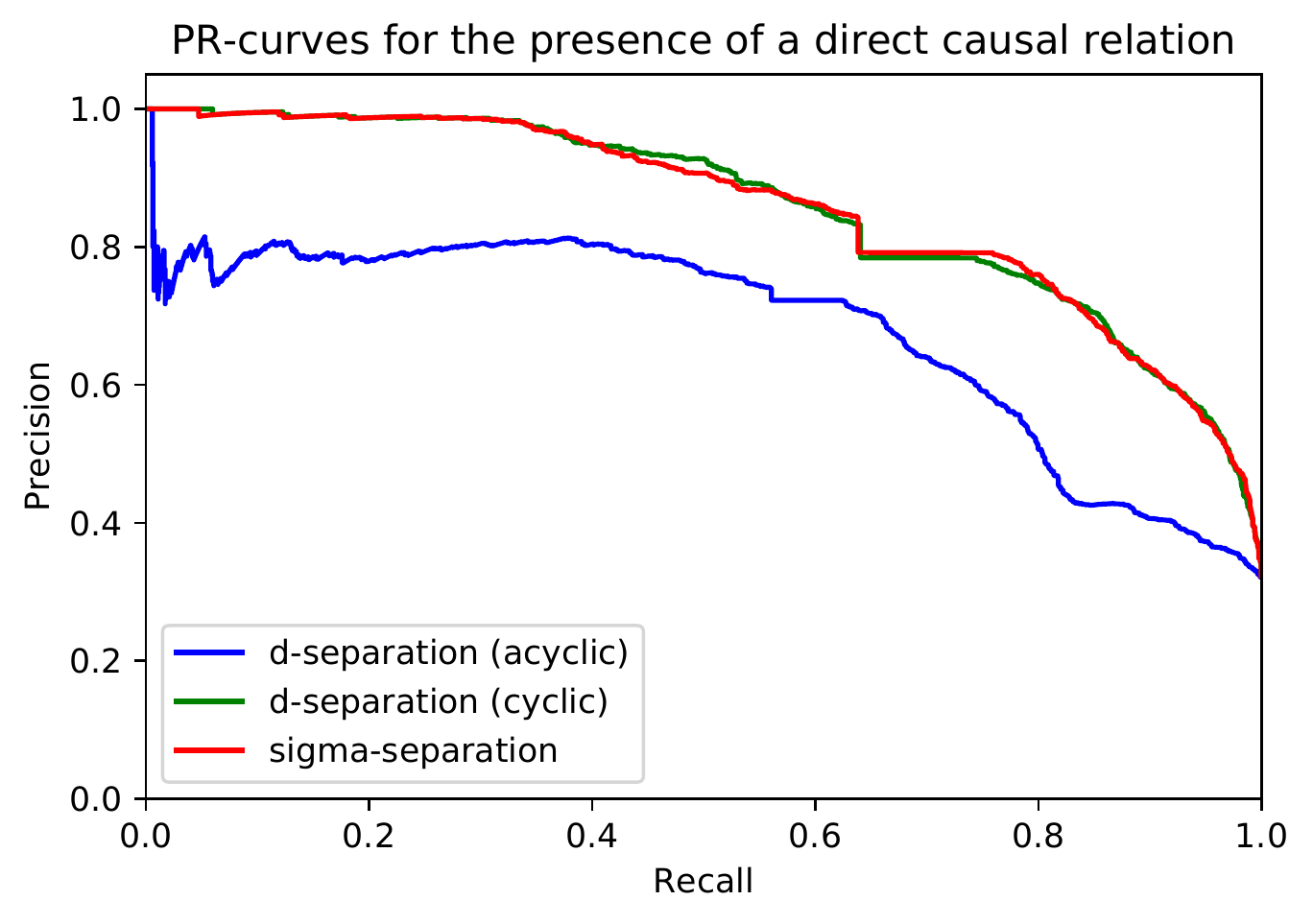}\qquad
	\includegraphics[scale=0.55]{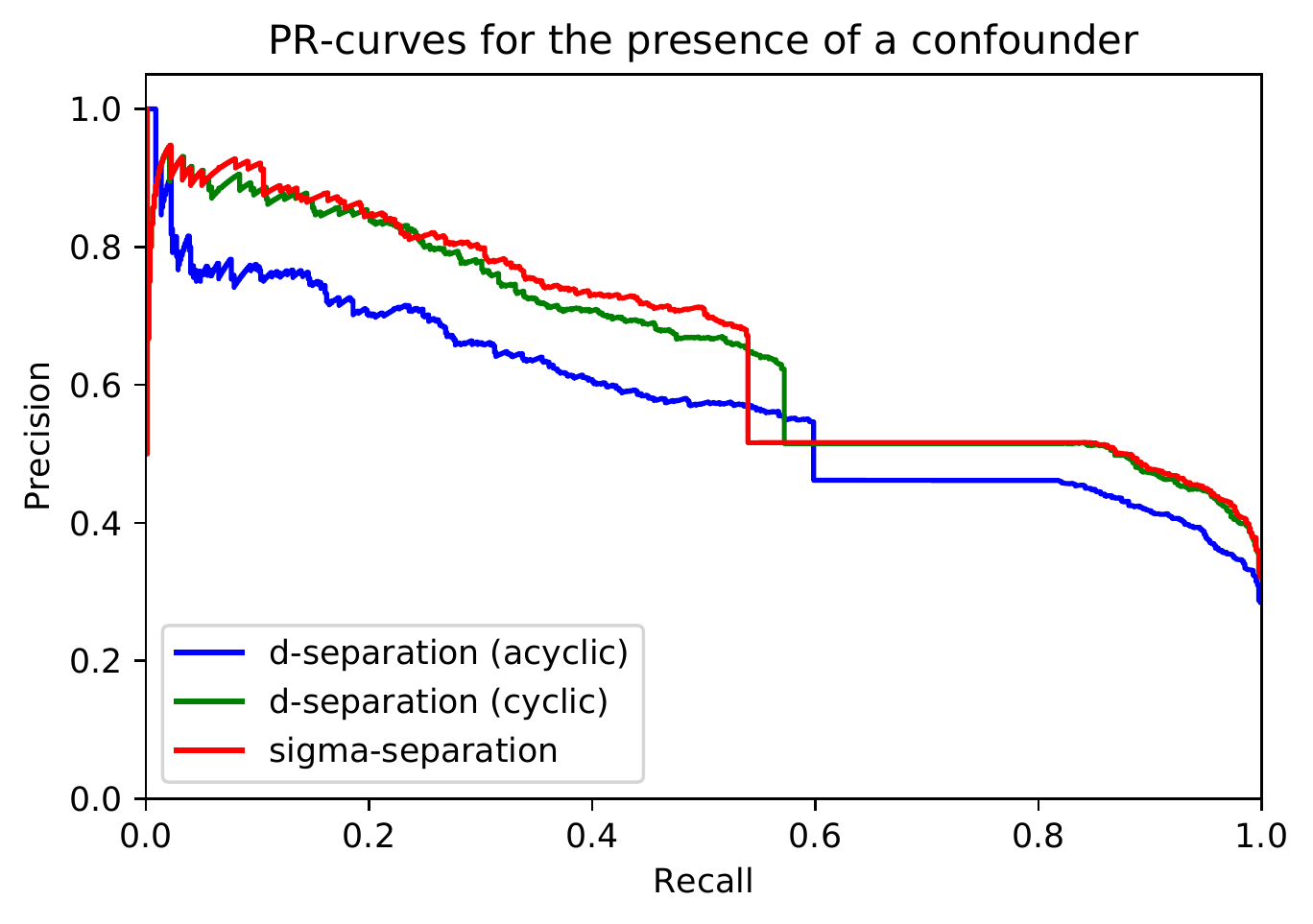}\\
  \\
	\includegraphics[scale=0.55]{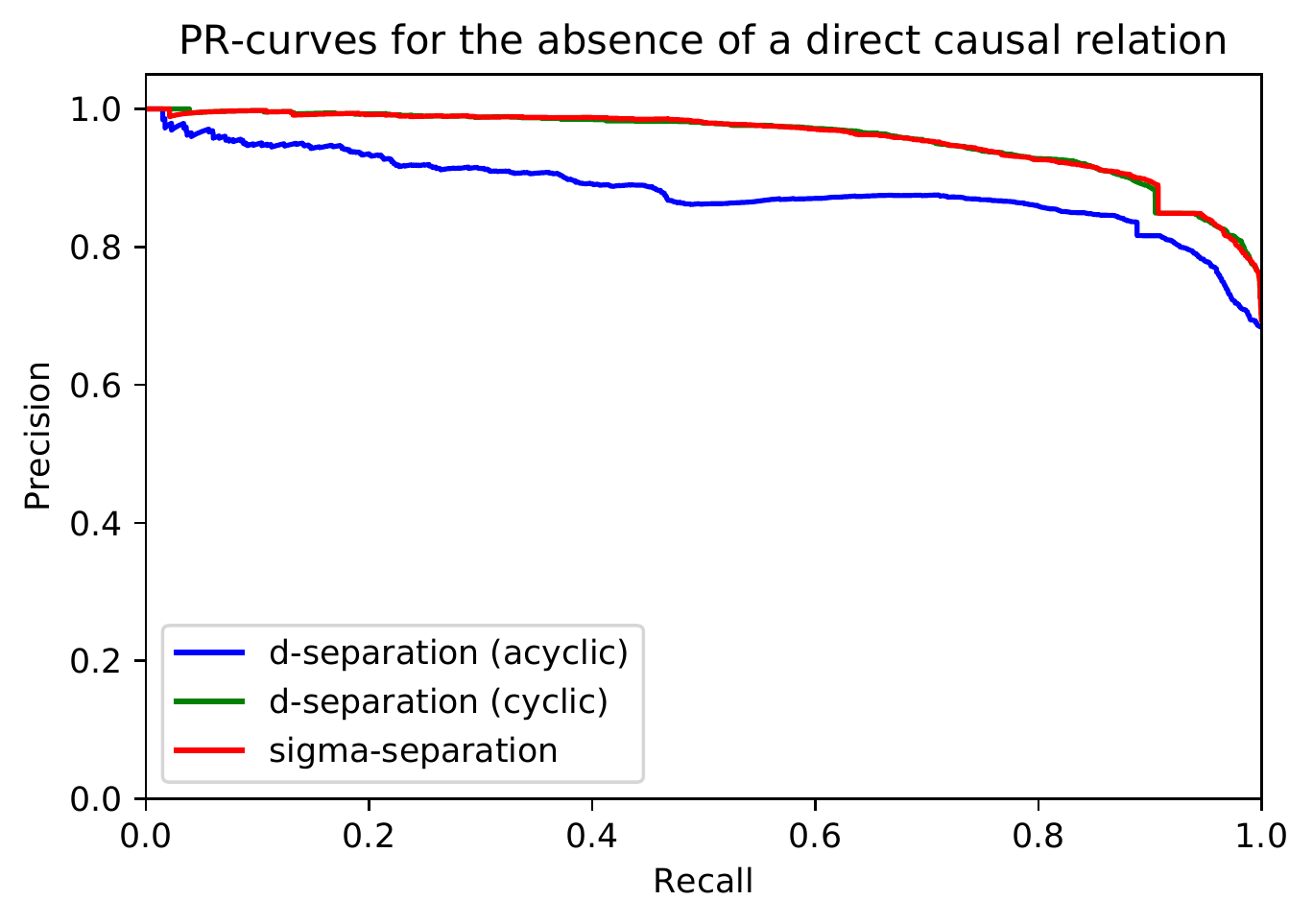}\qquad 
	\includegraphics[scale=0.55]{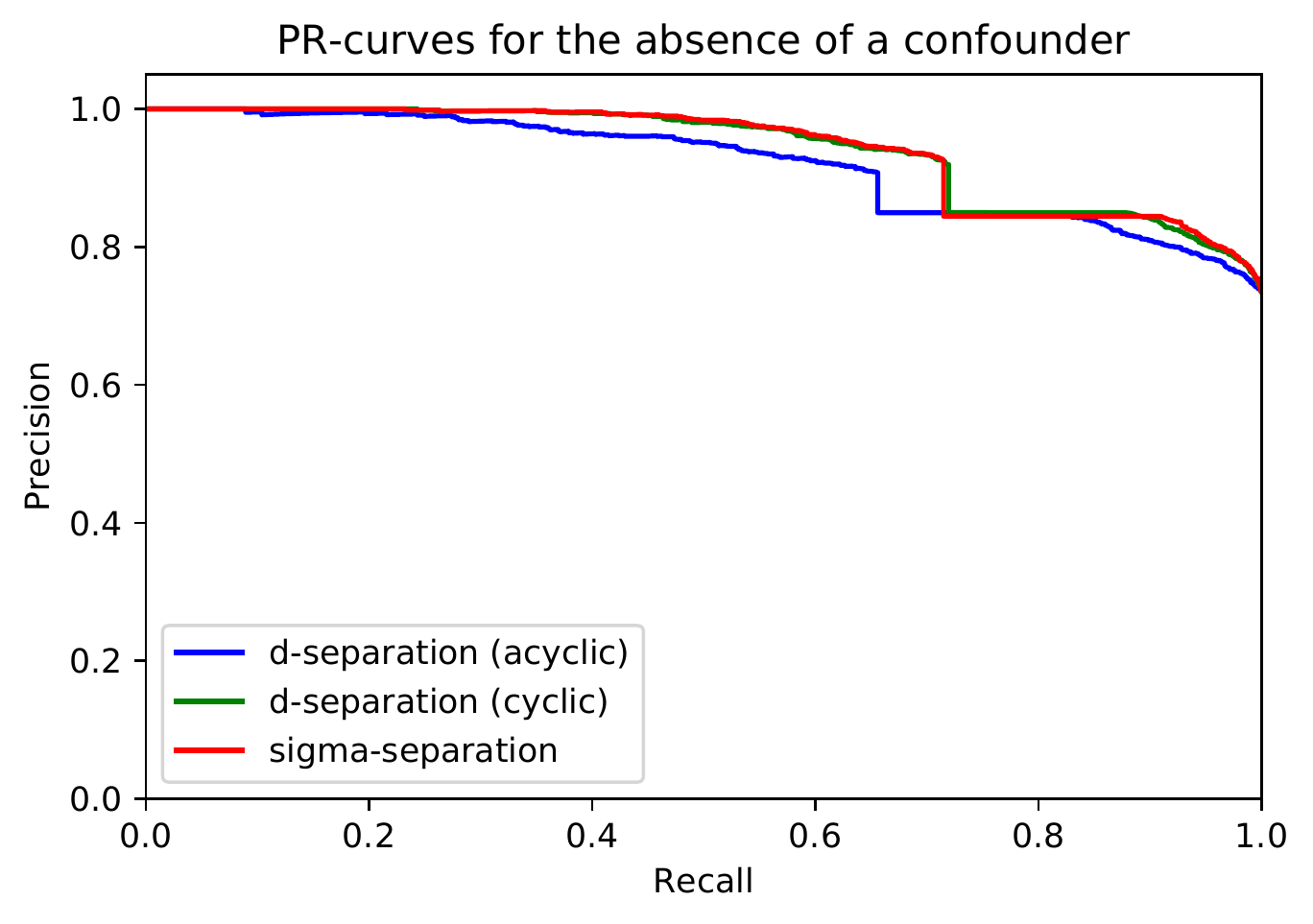}
  \caption{ROC curves (top) and PR curves (center, bottom) for directed edges (left) and confounders (right), for different encodings. All results shown here use observational and 5 interventional data sets.}
	 \label{fig:roc-pr-encoding}
\end{figure}

Figure~\ref{fig:roc-pr-encoding} shows similar curves, now for 5 interventions
only, but for different encodings: $\sigma$-separation (this work), d-separation
(allowing for cycles, \cite{HEJ14}) and d-separation (acyclic, \cite{HEJ14}).
Interestingly, the differences between $\sigma$-separation and 
d-separation turn out to be quite small in our simulation setting. The difference
is largest for the detection of confounders. On the other hand, the difference between
assuming acyclicity and allowing for cycles is much more pronounced, and is also
significant for the detection of direct causal relations. 

We expect that when
going to larger graphs with more variables and with nested loops, the differences between
$\sigma$-separation and d-separation should increase. However, due to computational restrictions 
we were not able to perform sufficiently many experiments in this regime to gather enough 
empirical support for that hypothesis and leave this for future research.

\onecolumn
\newpage

\end{document}